\documentclass{article}

\usepackage{microtype}
\usepackage{graphicx}
\usepackage{subfigure}
\usepackage{booktabs} % for professional tables
\usepackage{hyperref}
\usepackage[accepted]{icml2023}

\usepackage{amsmath}
\usepackage{amssymb}
\usepackage{mathtools}
\usepackage{amsthm}
\usepackage[ruled,vlined]{algorithm2e}
\usepackage{multirow}
\usepackage{xcolor,colortbl}

\usepackage[capitalize,noabbrev]{cleveref}
\crefname{section}{Sec.}{Secs.}
\Crefname{section}{Section}{Sections}
\Crefname{table}{Table}{Tables}
\crefname{table}{Tab.}{Tabs.}
\crefname{equation}{Eq.}{Eqs.}
\crefname{algorithm}{Alg.}{Algs.}
\crefname{figure}{Fig.}{Figs.}
\crefname{appendix}{App.}{Apps.}

\theoremstyle{plain}
\newtheorem{theorem}{Theorem}[section]

\theoremstyle{definition}
\newtheorem{definition}[theorem]{Definition}

\theoremstyle{remark}

\usepackage{enumitem}
\setenumerate[1]{itemsep=0pt,partopsep=0pt,parsep=\parskip,topsep=5pt}
\setitemize[1]{itemsep=0pt,partopsep=0pt,parsep=\parskip,topsep=5pt}
\setdescription{itemsep=0pt,partopsep=0pt,parsep=\parskip,topsep=5pt}

\usepackage[textsize=tiny]{todonotes}

\icmltitlerunning{Learning to Learn from APIs: Black-Box Data-Free Meta-Learning}

\begin{document}

\twocolumn[
\icmltitle{Learning to Learn from APIs: Black-Box Data-Free Meta-Learning}



\icmlsetsymbol{equal}{*}

\begin{icmlauthorlist}
\icmlauthor{Zixuan Hu}{sigs}
\icmlauthor{Li Shen}{jd}
\icmlauthor{Zhenyi Wang}{buf}
\icmlauthor{Baoyuan Wu}{cuhksz}
\icmlauthor{Chun Yuan}{sigs}
\icmlauthor{Dacheng Tao}{syd}
\end{icmlauthorlist}

\icmlaffiliation{sigs}{Tsinghua Shenzhen International Graduate School, Tsinghua University, Shenzhen, China}
\icmlaffiliation{jd}{JD Explore Academy, Beijing, China}
\icmlaffiliation{buf}{Department of Computer Science and Engineering, University at Buffalo, NY, USA}
\icmlaffiliation{cuhksz}{School of Data Science, the Chinese University of Hong Kong, Shenzhen, China}
\icmlaffiliation{syd}{School of Computer Science, the University of Sydney, Sydney, Australia}
\icmlcorrespondingauthor{Li Shen}{mathshenli@gmail.com}
\icmlcorrespondingauthor{Zhenyi Wang}{zhenyiwa@buffalo.edu}
\icmlcorrespondingauthor{Chun Yuan}{yuanc@sz.tsinghua.edu.cn}

\icmlkeywords{meta-learning, data-free, black-box}

\vskip 0.3in
]



\printAffiliationsAndNotice{} % otherwise use the standard text.

\begin{abstract}
Data-free meta-learning (DFML) aims to enable efficient learning of new tasks by meta-learning from a collection of pre-trained models without access to the training data.  
Existing DFML work can only meta-learn from (i) white-box and (ii) small-scale pre-trained models (iii) with the same architecture, neglecting the more practical setting where the users only have inference access to the APIs with arbitrary model architectures and model scale inside.
To solve this issue, we propose a \textbf{Bi}-level \textbf{D}ata-\textbf{f}ree \textbf{M}eta \textbf{K}nowledge \textbf{D}istillation (\textbf{BiDf-MKD}) framework to transfer more general meta knowledge from a collection of black-box APIs to one single meta model. 
Specifically, by just querying APIs, we inverse each API to recover its training data via a zero-order gradient estimator and then perform meta-learning via a novel bi-level meta knowledge distillation structure, in which we design a boundary query set recovery technique to recover a more informative query set near the decision boundary. 
In addition, to encourage better generalization within the setting of limited API budgets, we propose task memory replay to diversify the underlying task distribution by covering more interpolated tasks. Extensive experiments in various real-world scenarios show the superior performance of our BiDf-MKD framework. Code is available at \url{https://github.com/Egg-Hu/BiDf-MKD}.
\end{abstract}

\section{Introduction}
Data-free meta-learning (DFML) aims to meta-learn the useful prior knowledge from a collection of pre-trained models to enable efficient learning of new tasks without access to the training data due to privacy issues.
Existing DFML work \cite{wang2022metalearning} can only deal with the white-box pre-trained models, assuming access to the underlying model architecture and parameters. However, this assumption is not always satisfied. 
Recently, the concept of Model as A Sevice (MaaS) \cite{roman2009model} comes to reality.
Without access to the underlying models, users only have inference access to the corresponding APIs provided by the service providers like OpenAI or Google.
For example, Cloud Vision API of Google, Amazon AI and Alibaba Cloud provide thousands of APIs designed for solving various specific tasks. 
% ChatGPT provides a conversational query way to the users. 
TensorFlow Lite APIs provides numerous lightweight APIs deployed on mobiles, microcontrollers and edge devices.
In this paper, we argue that these APIs can not only be the black-box tools for solving specific tasks, but can also serve as the training resources of meta-learning to enable efficient learning of new unseen tasks. The significance of doing this is to remove the need for large volumes of labeled data to perform meta-learning and reliably protect data privacy and security. 
This motivation leads to our valuable but challenging topic, i.e., black-box DFML, which aims to meta-learn the meta-initialization from a collection of black-box APIs without access to the training data and with only inference access, to enable efficient learning of new tasks without data privacy leakage.

The main challenges of black-box DFML lie in three aspects: (i) \textit{data-free}: we have no access to the original training data of each API; (ii) \textit{black-box}: we have no prior knowledge of the underlying model architecture and parameters inside each API; (iii) \textit{model-agnostic}: each API may correspond to arbitrary underlying model architectures and model scale. Existing DFML work \cite{wang2022metalearning} only tries to handle the first challenge, which can not meta-learn from black-box APIs with arbitrary underlying model architecture and scale. Concretely, Wang et al. \yrcite{wang2022metalearning} propose to meta-learn a neural network to predict the meta-initialization given a collection of white-box pre-trained models. However, this method requires the exact parameter of each pre-trained model, and it requires all pre-trained models share the same architecture. Besides, it can not scale to large-scale pre-trained models because it directly uses a hyper neural network to output all parameters of the meta-initialization.

\begin{figure*}[t]
    \centering
    \includegraphics[width=0.95\linewidth]{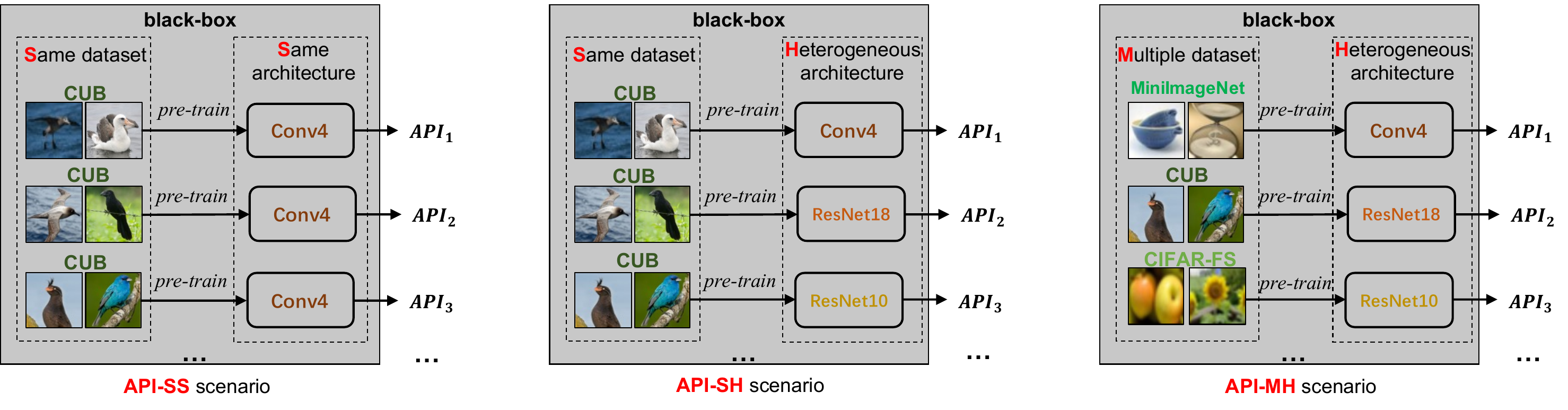}
    \vspace{-0.2cm}
    \caption{According to the datasets and model architectures inside the APIs, we propose three real-world black-box scenarios for a complete and practical evaluation of black-box DFML. We are the first to propose a unified framework simultaneously applicable to all three scenarios without any change, thus greatly expanding the real-world application scope of black-box DFML.
    }
    \vspace{-0.2cm}
    \label{fig:scenario}
\end{figure*}

In this work, we solve all these issues in a unified framework (see \cref{fig:pipeline}). We propose a novel \textbf{Bi}-level \textbf{D}ata-\textbf{f}ree \textbf{M}eta \textbf{K}nowledge \textbf{D}istillation (\textbf{BiDf-MKD}) framework to transfer more general meta knowledge from a collection of black-box APIs to one single meta model, which serves as the meta-initialization to initialize the task-specific models of new unseen tasks. 
Specifically, by just querying the API, we first ``inverse" each API to recover the label-conditional data starting from latent standard Gaussian noise via a zero-order gradient estimator. With the aid of the recovered data, we then perform meta-learning by transferring the general meta knowledge from a collection of black-box APIs to the meta model through our proposed bi-level meta knowledge distillation structure. 
Then, we formally define the \textit{knowledge vanish} (see \cref{def:inj}) issue involved in the bi-level characteristic of meta-learning from the perspective of information theory, i.e., the outer-level optimization of meta-learning could be useless. 
We argue that the knowledge vanish issue in the data-free setting is more significant than in the data-based setting due to the relatively low diversity of recovered data.
As illustrated in \cref{fig:boundary}, to alleviate such issue, we design a boundary query set recovery technique to amplify the diversity by recovering a more informative query set near the decision boundary.
In addition, to encourage better generalization to the unseen tasks within the setting of limited API budgets, we propose task memory replay on more interpolated tasks. The interpolated tasks do not correspond to any API so that we can diversify the underlying task distribution associated with the given APIs by covering more new tasks. 
Overall, our proposed framework can effectively solve the black-box DFML problem (i) without the need for real data, (ii) with only inference access to the APIs, (iii) regardless of the underlying model architecture and model scale inside each API, and (iv) without data privacy leakage. Thus, it substantially expands the real-world application scenarios of black-box DFML.

We perform extensive experiments in three real-world black-box scenarios (see \cref{fig:scenario}), including (i) \textbf{API-SS}. All APIs are designed for solving tasks from the \textbf{S}ame meta training subset with the \textbf{S}ame architecture inside. (ii) \textbf{API-SH}. All APIs are designed for solving tasks from the \textbf{S}ame meta training subset but with \textbf{H}eterogeneous architectures inside. (iii) \textbf{API-MH}. All APIs are designed for solving tasks from \textbf{M}ultiple meta training subsets with \textbf{H}eterogeneous architectures inside. 
For benchmarks of three scenarios on CIFAR-FS, MiniImageNet and CUB, our framework achieves significant performance gains in the range of $8.09\%$ to $21.46\%$.
We summarize the main contributions as three-fold:
\begin{itemize}

    \item For the first time, we propose a new practical and valuable setting of DFML, i.e., black-box DFML, whose goal is to meta-learn the meta-initialization from a collection of black-box APIs without access to the original training data and with only inference access, to enable efficient learning of new tasks without privacy leakage.

    \item We propose BiDf-MKD to meta-learn the meta-initialization by transferring general meta knowledge from a collection of black-box APIs to one single model. We formally define the knowledge vanish issue of meta-learning and design the boundary query set recovery technique to alleviate it. We also propose task memory replay to boost the generalization ability for the setting of limited API budgets.
    
    \item We propose three real-world black-box scenarios (API-SS, API-SH, and API-MH) for a complete and practical evaluation of black-box DFML. We are the first to propose a data-free, inference-based and model-agnostic framework, simultaneously applicable to all three scenarios without any change and outperforming the SOTA baselines by a large margin.
\end{itemize}

\section{Related Work}
\label{related_work}

\textbf{Meta-learning \& Data-free meta-learning.}\ Meta-learning \cite{schmidhuber1987evolutionary}, a.k.a. \textit{learning to learn}, aims to meta-learn useful prior knowledge from a collection of tasks, which can be generalized to new unseen tasks efficiently. 
MAML \cite{finn2017model} and its variants \cite{abbas2022sharp,jeong2020ood,raghu2019rapid,behl2019alpha,rajeswaran2019meta} meta-learn a sensitive meta-initialization to initialize the task-specific model, while other works \cite{santoro2016meta,mishra2017simple,garnelo2018conditional,munkhdalai2017meta} meta-learn a hyper neural network to output the task-specific model parameters conditioned on the support set. 
Existing meta-learning works \cite{ vinyals2016matching,wang2021meta,finn2018probabilistic,yao2021meta,wang2022meta,harrison2020continuous,ye2020few,li2020boosting,yang2021free,simon2022meta,liu2019learning,Wang_2022_CVPR,wang2020bayesian,zhou2021meta} assume the access to the training data associated with each task. More recently, Wang et al. \yrcite{wang2022metalearning} propose a new meta-learning paradigm, data-free meta-learning, which aims to meta-learn the meta-initialization from a collection of pre-trained modes without access to their training data. However, it imposes strict restrictions on pre-trained models: (i) white-box, (ii) small-scale, and (iii) with the same architecture, thus reducing its applicable scenarios in real applications. \nocite{hu2023architecture}

\textbf{Knowledge distillation for meta-learning.}\ Our work is reminiscent of knowledge distillation (KD) \cite{44873} as we leverage a collection of APIs to supervise the meta-learning. Below, we first briefly review KD and compare ours with existing KD works for meta-learning. 
KD aims to supervise the training process of the student model with the knowledge of the teacher model. The knowledge can be the soft-label predictions \cite{44873}, hidden layer activation \cite{romero2014fitnets,AnimeshKoratana2019LITLI}, embedding \cite{HantingChen2018LearningSN,SungsooAhn2019VariationalID}, or relationship \cite{HanJiaYe2020DistillingCK}. Existing meta-learning works based on KD differ greatly from ours in motivation, setting and manner. Ye et al. \yrcite{ye2022few} associate each task with an additional teacher classifier to provide additional supervision for meta-learning. They conduct KD by minimizing the prediction disagreement on real data, thus not applicable to the data-free setting. Besides, it relies on separate training with real data to obtain the teacher classifiers. REFILLED \cite{HanJiaYe2020DistillingCK} performs KD between one teacher model and one student model in different label spaces. It relies on the relationship among embeddings, which are unavailable in our black-box setting. 

\textbf{Model inversion.}\ Model inversion \cite{fredrikson2015model,wu2016methodology,zhang2020secret} aims to recover the training data from the pre-trained model. Our framework also involves recovering data from black-box APIs to transfer meta knowledge. Existing techniques \cite{fredrikson2015model,wu2016methodology,zhang2020secret,deng2021graph,lopes2017data,chawla2021data,zhu2021data,liu2021data,zhang2022fine,fang2021contrastive} about model inversion from white-box pre-trained models are not applicable to our black-box setting. Recent works DFME \cite{truong2021data} and MAZE \cite{kariyappa2021maze} leverage the black-box model inversion technique to perform model extraction. Key differences include our meta-learning objective and loss formulation for label-conditional data recovery. We also design a novel boundary data recovery technique to recover more informative data near the decision boundary (see \cref{fig:boundary}).
\section{Problem Setup}
\label{sec:problem-setup}

In this section, we first clarify the definition of black-box DFML, followed by its meta testing procedure.

\subsection{Black-box DFML Setup}

We are given a collection of APIs $\{A_i\}$ solving different tasks, with only inference access and without accessing their original training data. We aim to meta-learn the meta-initialization $\boldsymbol{\theta}$, which can be adapted fast to new unseen tasks $\{\mathcal{T}_{i}^{new}\}$.
Note that each API may correspond to arbitrary underlying model architecture and model scale. 

\subsection{Meta Testing}

During meta testing, several unseen $N$-way $K$-shot tasks $\{\mathcal{T}_{i}^{new}=\{\boldsymbol{S}_{i}^{new},\boldsymbol{Q}_{i}^{new}\}\}$ arrive together. The classes appearing in meta testing tasks are unseen during meta training. Each task contains a support set $\boldsymbol{S}_{i}^{new}$ with $N$ classes and $K$ instances per class. 
We use the support set $\boldsymbol{S}_{i}^{new}$ to adapt the meta initialization to the task-specific task (i.e., $\boldsymbol{\theta} \rightarrow \boldsymbol{\theta}_i^{new}$). The query set $\boldsymbol{Q}_{i}^{new}$ is what we actually need to predict. 
The final accuracy is measured by the average accuracy for those meta testing tasks.

\section{Methodology}
\label{methodology}
\begin{figure*}[t]
    \centering
    \includegraphics[width=0.8\linewidth]{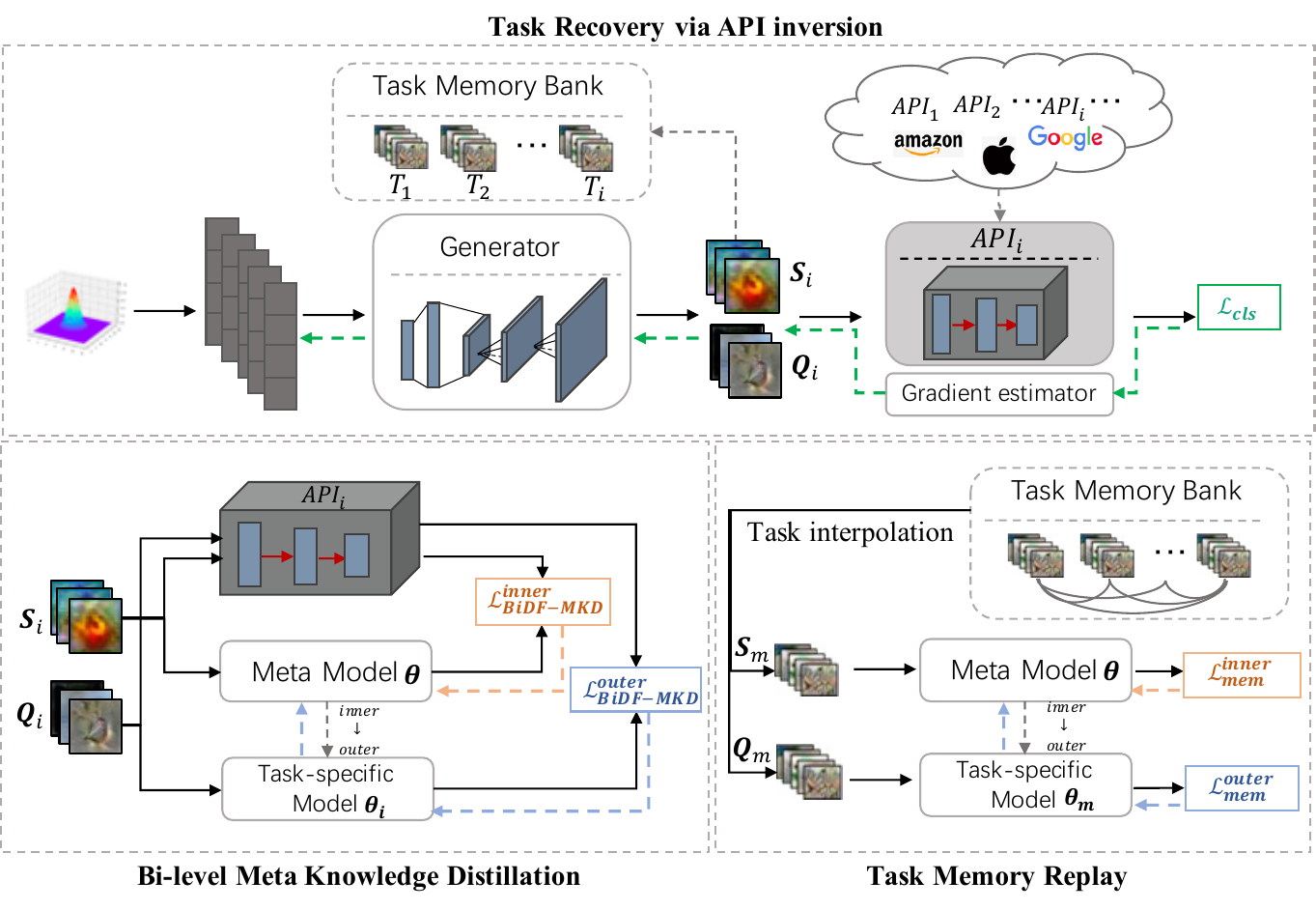}
    \vspace{-0.2cm}
    \caption{The whole pipeline of our proposed BiDf-MKD framework. For each API $A_i$, we recover its training data starting from the random standard Gaussian noise $\boldsymbol{Z}_i$. By continually querying the black-box API $A_i$, we gradually update the noise to label-conditional data. We then split the recovered data into the support set $\boldsymbol{S}_i$ and query set $\boldsymbol{Q}_i$ to perform meta-learning via our bi-level meta knowledge distillation structure. Alternatively, we can perform task memory replay with MAML over more interpolated tasks. 
    % Here, for brevity, we omit the boundary query set recovery technique for recovering a more informative query set.
    }
    \vspace{-0.2cm}
    \label{fig:pipeline}
\end{figure*}

In this section, we propose a unified framework (\cref{fig:pipeline}) to solve the black-box DFML problem, including (i) BiDf-MKD  to transfer meta knowledge (\cref{sub:bidfkd}) and (ii) task memory replay to boost generalization ability (\cref{sub:memory}).

\subsection{Bi-level data-free meta knowledge distillation (BiDf-MKD)}
\label{sub:bidfkd}

\textbf{Task recovery via API inversion.}\ For the API $A_{i}$, we aim to recover the task-specific training data set $\hat{\boldsymbol{X}}$, with which general meta knowledge can be transferred from the API $A_{i}(\cdot)$ to the meta model $F(\cdot; \boldsymbol{\theta})$. As illustrated in \cref{fig:pipeline}, it consists of four components: a generator $G(\cdot; \boldsymbol{\theta}_{G})$, an API $A_{i}$, a gradient estimator and a memory bank $\mathcal{B}$. The generator $G(\cdot; \boldsymbol{\theta}_{G})$ takes the standard Gaussian noise $\boldsymbol{z}$ as input and outputs the recovered data $\hat{\boldsymbol{x}}=G(\boldsymbol{z}; \boldsymbol{\theta}_G)$. To make the recovery process label-conditional, we update the corresponding $\boldsymbol{z}$ and $\boldsymbol{\theta}_{G}$ simultaneously by minimizing the per datum cross-entropy loss
\begin{equation}
\small
    \min_{\boldsymbol{z},\boldsymbol{\theta}_G} \ {\ell}_{cls}(\hat{\boldsymbol{x}},y)=CE(A_{i}(\hat{\boldsymbol{x}}),y),
    \ \ \ \rm{s.t.} \ \ \hat{\boldsymbol{x}}=G(\boldsymbol{z}; \boldsymbol{\theta}_G),
    \label{eq:losscls}
\end{equation}
where $y$ is the pre-defined class label.
To recover a batch of data $\hat{\boldsymbol{X}}$ of class labels $\boldsymbol{Y}$, we update $\boldsymbol{Z}$ and $\boldsymbol{\theta}_{G}$ by minimizing the batch-wise loss 
\begin{equation}
\small
\begin{split}
\small
    &\min_{\boldsymbol{Z},\boldsymbol{\theta}_G} \ \mathcal{L}_{cls}(\hat{\boldsymbol{X}})=\frac{1}{|\hat{\boldsymbol{X}}|}\sum_{(\hat{\boldsymbol{x}},y) \in (\hat{\boldsymbol{X}},\boldsymbol{Y})}l_{cls}(\hat{\boldsymbol{x}},y),\\
    &\quad \rm{s.t.} \ \ \hat{\boldsymbol{X}}=G(\boldsymbol{Z}; \boldsymbol{\theta}_G).
    \label{eq:lossclsbatch}
    \end{split}
\end{equation}
After obtaining a certain number of data, we feed these recovered $\hat{\boldsymbol{X}}$ into the memory bank $\mathcal{B}$, i.e., a first-in-first-out (FIFO) container with a certain volume.

\textbf{Zero-order gradient estimation.}\ Recall that our objective of task recovery is to update $\boldsymbol{z} \in \boldsymbol{Z}$ and $\boldsymbol{\theta}_{G}$ simultaneously by minimizing $\mathcal{L}_{cls}$.
\begin{subequations}
\small
    \begin{align}
    \boldsymbol{\theta}_{G}^{t+1}=\boldsymbol{\theta}_{G}^{t}-\eta \nabla_{\boldsymbol{\boldsymbol{\theta}_{G}}}\mathcal{L}_{cls}\\
    \boldsymbol{z}^{t+1}=\boldsymbol{z}^{t}-\eta \nabla_{\boldsymbol{z}}\mathcal{L}_{cls}.
    \end{align}
    \label{eq:update}
\end{subequations}
Updating $\boldsymbol{z}$ and $\boldsymbol{\theta}_{G}$ in such way involves calculating $\nabla_{\boldsymbol{\boldsymbol{\theta}_{G}}}\mathcal{L}_{cls}$ and $\nabla_{\boldsymbol{z}}\mathcal{L}_{cls}$. With the use of the chain rule, we decompose each gradient into two components: 
% (with $\mathcal{L}=\mathcal{L}_{cls}$ and $\ell=\ell_{cls}$):
\begin{subequations}
\small
    \begin{align}
    \label{eq:decompose1}
    \nabla_{\boldsymbol{\boldsymbol{\theta}_{G}}}\mathcal{L}_{cls}&=\frac{\partial \mathcal{L}_{cls}}{\partial \boldsymbol{\theta}_G}=\frac{1}{|\hat{\boldsymbol{X}}|}\sum_{\hat{\boldsymbol{x}} \in \hat{\boldsymbol{X}}}\left[\frac{\partial \ell_{cls}}{\partial \hat{\boldsymbol{x}}}\times\frac{\partial \hat{\boldsymbol{x}}}{\partial \boldsymbol{\theta}_G}\right]\\
    &\nabla_{\boldsymbol{z}}\mathcal{L}_{cls}=\frac{\partial \mathcal{L}_{cls}}{\partial \boldsymbol{z}}=\frac{\partial {\ell_{cls}}}{\partial \hat{\boldsymbol{x}}}\times\frac{\partial \hat{\boldsymbol{x}}}{\partial \boldsymbol{z}}.
    \label{eq:decompose2}
    \end{align}
    \label{eq:decompose}
\end{subequations}
The second factors ($\frac{\partial \hat{\boldsymbol{x}}}{\partial \boldsymbol{\theta}_G}$ and $\frac{\partial \hat{\boldsymbol{x}}}{\partial \boldsymbol{z}}$) in \cref{eq:decompose1} and \cref{eq:decompose2} can be automatically calculated via the automatic differentiation mechanism in PyTorch \cite{paszke2017automatic} or TensorFlow \cite{tensorflow2015-whitepaper}. However, it is not applicable for calculating the first factor ($\frac{\partial {\ell_{cls}}}{\partial \hat{\boldsymbol{x}}}$), because we have no access to the underlying model parameters inside the API. To this end, we adopt a zero-order gradient estimator to obtain an approximation of the first-order gradient by just querying the API. To explain how the first-order gradient ($\frac{\partial {\ell_{cls}}}{\partial \hat{\boldsymbol{x}}}$) is estimated, consider the random noise vector $\boldsymbol{z} \in \mathbb{R}^{d_{\boldsymbol{z}}}$, which yields the recovered data $G(\boldsymbol{z};\boldsymbol{\theta}_G)=\hat{\boldsymbol{x}} \in \mathbb{R}^{d_{\boldsymbol{\hat{\boldsymbol{x}}}}}$ with label $y$ and the loss value $\ell_{cls}(\hat{\boldsymbol{x}},y) \in \mathbb{R}$. 
We can feed the recovered data plus a set of random direction vectors to query the API and estimate the gradient according to the difference of two loss values. This leads to the randomized gradient estimation \cite{liu2020primer}:
\begin{equation}
\small
    \hat{\nabla}_{\hat{\boldsymbol{x}}} \ell_{cls}\!=\!\frac{1}{q} \sum_{i=1}^q\left[\frac{d_{\hat{\boldsymbol{x}}}}{\mu}\left(\ell_{cls}\left(\hat{\boldsymbol{x}}+\mu \boldsymbol{u}_i,y\right)\!-\ell_{cls}(\hat{\boldsymbol{x}},y)\right) \boldsymbol{u}_i\right],
    \label{eq:estimation}
\end{equation}
where $\{\boldsymbol{u}_i\}_{i=1}^q$ are $q$ random direction vectors sampled independently and uniformly from the sphere of a unit ball. $\mu > 0$, a.k.a the smoothing parameter, is a given small step size. 
The estimation $\hat{\nabla}_{\hat{\boldsymbol{x}}} \ell_{cls}$ is reasonable because \cref{eq:estimation} provides an unbiased estimate of the first-order gradient ${\nabla}_{\hat{\boldsymbol{x}}} \ell_{cls}$ of the Gaussian smoothing version \cite{gao2018information,zhang2022how}.
With the zero-order gradient estimator, we can obtain the estimated gradient according to \cref{eq:decompose}:
\begin{subequations}
\small
    \begin{align}
    \nabla_{\boldsymbol{\boldsymbol{\theta}_{G}}}\mathcal{L}_{cls}&\approx \frac{1}{|\hat{\boldsymbol{X}}|}\sum_{\hat{\boldsymbol{x}} \in \hat{\boldsymbol{X}}}\left[\hat{\nabla}_{\hat{\boldsymbol{x}}} \ell_{cls}\times\frac{\partial \hat{\boldsymbol{x}}}{\partial \boldsymbol{\theta}_G}\right]\\
    &\nabla_{\boldsymbol{z}}\mathcal{L}_{cls}\approx \hat{\nabla}_{\hat{\boldsymbol{x}}} \ell_{cls}\times\frac{\partial \hat{\boldsymbol{x}}}{\partial \boldsymbol{z}},
    \end{align}
\end{subequations}
where $\frac{\partial \hat{\boldsymbol{x}}}{\partial \boldsymbol{\theta}_G} \in \mathbb{R}^{d_{\hat{\boldsymbol{x}}}\times d_{\boldsymbol{\theta}_{G}}}$ and $\frac{\partial \hat{\boldsymbol{x}}}{\partial \boldsymbol{z}} \in \mathbb{R}^{d_{\hat{\boldsymbol{x}}}\times d_{\boldsymbol{z}}}$ are the Jacobian matrices and $\hat{\nabla}_{\hat{\boldsymbol{x}}} \ell_{cls} \in \mathbb{R}^{1\times d_{\hat{\boldsymbol{x}}}}$ is the zero-order gradient estimation. Then we can perform gradient descent by updating $\boldsymbol{Z}$ and $\boldsymbol{\theta}_{G}$ according to \cref{eq:update} to produce the recovered data required to perform meta knowledge distillation.

\textbf{Bi-level meta knowledge distillation for meta-learning.}\ We propose a bi-level meta knowledge distillation structure to perform meta-learning by transferring general meta knowledge from a collection of black-box APIs into one single meta model with the recovered data. Different from the common knowledge distillation methods requiring the teacher and student designed for the same task, the meta model is not tailored to any specific task. Thus, it is not appropriate to directly transfer the task-specific knowledge from the API (viewed as the teacher) to the meta model (viewed as the student). 
To this end, our bi-level structure controls the knowledge flow from each API to the meta model via an intermediate task-specific model, which transfers more general meta knowledge. 
The meta knowledge enables fast knowledge distillation of the inner loop, thus not task-specific and approximate to be transferred to the meta model.

Our proposed BiDf-MKD involves a bi-level structure, i.e., the inner level and the outer level. For API $A_i$, we split its recovered data $\hat{\boldsymbol{X}}_i$ into two non-overlap support set $\boldsymbol{S}_i$ and query set $\boldsymbol{Q}_i$. For the inner level, we transfer the task-specific knowledge from $A_i$ to a task-specific model $F(\cdot; \boldsymbol{\theta}_i)$ initialized by $\boldsymbol{\theta}$ so that the task-specific model $F(\cdot; \boldsymbol{\theta}_i)$ can act like $A_i$ on $\boldsymbol{S}_i$. We clone this task-specific model $F(\cdot; \boldsymbol{\theta}_i)$ from API $A_i$ by minimizing the disagreement of predictions between them:
\begin{equation}
\small
    \!\boldsymbol{\theta}_i\!=\min_{\theta}\mathcal{L}_{BiDf-MKD}^{inner} \! \triangleq \! \min_{\boldsymbol{\theta}}\sum_{\hat{\boldsymbol{x}} \in \boldsymbol{S}_i}\ell_{KL}(F(\hat{\boldsymbol{x}};\boldsymbol{\theta}),A_i(\hat{\boldsymbol{x}})),
    \label{eq:inner}
\end{equation}
where $\ell_{KL}(p,q)$ measures the Kullback–Leibler (KL) divergence \cite{mackay2003information} between distributions $p$ and $q$. One can apply other measures that can characterize the difference of distributions.
The objective of the inner level is only to transfer the task-specific knowledge from the API $A_i$ to the task-specific model $F(\cdot;\boldsymbol{\theta}_i)$. Note that the task-specific knowledge is not desired for the meta model, because the meta model should possess the potential to work well with all tasks after adaptation and one certain task-specific knowledge can not directly adapt to other different tasks. Then, with the aid of the inner level, we resort to the outer level to explore more general meta knowledge, which is beneficial to the meta model.

The core idea of the outer level is to explore more general meta knowledge, with which we can make the best use of the task-specific knowledge in the inner level. In other words, it is hard to obtain an excellent task-specific model relying solely on the task-specific knowledge in the inner level with few recovered data. We desire more general meta knowledge to facilitate the inner-level knowledge distillation so as to narrow the gap between the task-specific model $F(\cdot;\boldsymbol{\theta}_i)$ and the API $A_i$ as much as possible. The gap is evaluated by testing the task-specific model $F(\cdot;\boldsymbol{\theta}_i)$ on a wider range of hold-out data $\boldsymbol{Q}_i$, which is equivalent to minimizing
\begin{equation}
\small
\begin{split}
        &\min_{\boldsymbol{\theta}}\ \mathcal{L}_{BiDf-MKD}^{outer}(\boldsymbol{\theta})=\sum_{\hat{\boldsymbol{x}} \in \boldsymbol{Q}_i}\ell_{KL}(F(\hat{\boldsymbol{x}};\boldsymbol{\theta}_i),A_i(\hat{\boldsymbol{x}})),\\
    \label{eq:outer}
     &\quad \rm{s.t. } \ \ 
    \boldsymbol{\theta}_i=\min_{\theta}\mathcal{L}_{BiDf-MKD}^{inner},
    % \triangleq \min_{\theta}\sum_{\hat{\boldsymbol{x}} \in \boldsymbol{S}_i}\ell_{KL}(F(\hat{\boldsymbol{x}};\boldsymbol{\theta}),A_i(\hat{\boldsymbol{x}}))
    \qquad\qquad\qquad
    \end{split}
\end{equation}
where $\boldsymbol{\theta}_i$ is obtained after the inner level following \cref{eq:inner} and $\boldsymbol{\theta}$ is the meta model parameters we truly update. $F(\hat{\boldsymbol{x}};\boldsymbol{\theta}_i)$ and $A_i(\hat{\boldsymbol{x}})$ outputs the prediction (after softmax) on $\hat{\boldsymbol{x}}$ from the task-specific model and API, respectively.

% Through the bi-level structure, we can control the knowledge flow from the API to the meta model via an intermediate task-specific model, which encourages the exploration of more general meta knowledge beneficial to the meta model.

\begin{figure}
    \centering
    \includegraphics[width=0.7\linewidth]{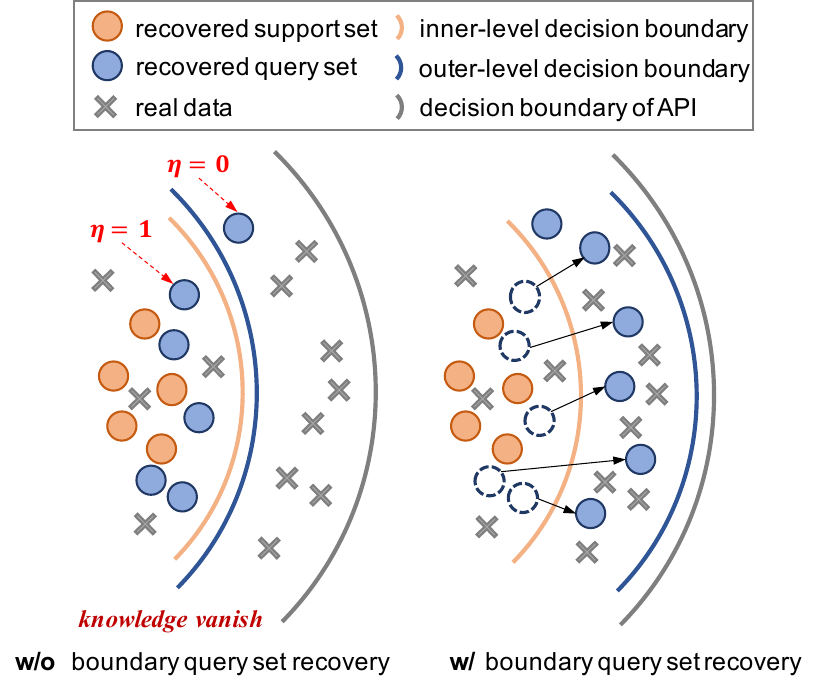}
    \vspace{-0.2cm}
    \caption{Knowledge vanish issue of meta-learning occurs when the outer-level optimization can be ignored.
    }
    \vspace{-0.2cm}
    \label{fig:boundary}
\end{figure}

\textbf{Boundary query set recovery for knowledge vanish.}\ 
The ideal BiDf-MKD is conducted in such a way that narrows the gap between the task-specific model and the API on the hold-out query set. However, the outer-level knowledge distillation could be too ``lazy" to explore the meta knowledge. This issue is more significant in the data-free meta-learning setting than in the data-based meta-learning setting because of the relatively low diversity of the recovered data. 
Consider an extreme example with $\boldsymbol{S}_i = \boldsymbol{Q}_i$. The task-specific model distilled from the API on $\boldsymbol{S}_i$ in the inner level can perform perfectly on $\boldsymbol{Q}_i$ in the outer level. This could lead to an illusion where the task-specific model is so ``perfect" that we can explore no more knowledge to facilitate the inner-level knowledge distillation. We then formally define the \textit{complete knowledge vanish} issue of meta-learning from the perspective of information theory.

\begin{definition}
\label{def:inj}
The \textit{complete knowledge vanish} of meta-learning occurs when the outer-level optimization can be ignored, namely the mutual information ${I}(\boldsymbol{\theta};\boldsymbol{Q}_i|\boldsymbol{\theta}_i,\boldsymbol{S}_i)=0$ (or ${H}(\boldsymbol{\theta}|\boldsymbol{\theta}_i,\boldsymbol{S}_i)=H(\boldsymbol{\theta}|\boldsymbol{\theta}_i,\boldsymbol{S}_i,\boldsymbol{Q}_i)$).
\end{definition}

Refer to \cref{app:mutualInformation} for the relation between the mutual information $I$ and entropy $H$. 
Explicitly maximizing ${I}(\boldsymbol{\theta};\boldsymbol{Q}_i|\boldsymbol{\theta}_i)$ requires an unknown posterior distribution over $\boldsymbol{\theta}$. Instead, we implicitly encourage ${H}(\boldsymbol{\theta}|\boldsymbol{\theta}_i,\boldsymbol{S}_i)>H(\boldsymbol{\theta}|\boldsymbol{\theta}_i,\boldsymbol{S}_i,\boldsymbol{Q}_i)$ by recovering $\boldsymbol{Q}_i$ with more information. Zhang et al. \yrcite{zhangdense} point out the samples near the decision boundary contained more valuable information for classification. 
This motivation leads to our proposed boundary query set recovery technique, which urges the generator to recover the query set between the decision boundaries of the task-specific model $F(\cdot;\boldsymbol{\theta}_{i})$ and API $A_{i}(\cdot)$ (see \cref{fig:boundary}). 
Specifically, we first recover the support set $\boldsymbol{S}_i$ (orange circles) by minimizing \cref{eq:lossclsbatch}. Then, we use $\boldsymbol{S}_i$ to conduct the inner-level knowledge distillation following \cref{eq:inner} to distill the task-specific model parameters $\boldsymbol{\theta}_i$ with the inner-level decision boundary (orange arc). To recover a more informative query set (blue circles) for the outer-level knowledge distillation, we incorporate $F(\cdot|\boldsymbol{\theta}_i)$ to query set recovery by maximizing the disagreement between $F(\cdot|\boldsymbol{\theta}_i)$ and the API $A_i(\cdot)$ following \cref{eq:adaptive}. Note that  large disagreement may guide to generate just some outliers. Therefore, we only pay more attention to those boundary samples and introduce the loss for boundary query set recovery:
\begin{equation}
\small
\begin{split}
    &\!\! \min_{\boldsymbol{z}, \boldsymbol{\theta}_G}\   {\ell}_{Q}(\hat{\boldsymbol{x}},y) \\
          &\quad \quad \!=\! CE(A_{i}(\hat{\boldsymbol{x}}),y) \!-\!\lambda_{Q}\cdot \eta \cdot \ell_{KL}(F(\hat{\boldsymbol{x}};\boldsymbol{\theta}_i),A_i(\hat{\boldsymbol{x}})),\\
          &\rm{s.t.} \ \ \hat{\boldsymbol{x}}=G(\boldsymbol{z}; \boldsymbol{\theta}_G),\\
          &\quad \quad  \eta=\mathbb{I}\{\arg \max F(\hat{\boldsymbol{x}};\boldsymbol{\theta}_i) = \arg \max A_i(\hat{\boldsymbol{x}}) \}.
          \label{eq:adaptive}
\end{split}
 \end{equation}
 The function $\mathbb{I}(\cdot)$ is an indicator to enable $\hat{\boldsymbol{x}}$ with the same prediction from the API and the task-specific model ($\eta = 1$), otherwise disable it ($\eta = 0$). Unlike the loss \cref{eq:losscls}, the loss \cref{eq:adaptive} guides to recover $\hat{\boldsymbol{x}}$ between the decision boundaries of the task-specific model $F(\cdot;\boldsymbol{\theta}_{i})$ (orange arc) and API $A_{i}(\cdot)$ (grey arc), which provides more information for the outer-level knowledge distillation.

\subsection{Task memory replay}
\label{sub:memory}
The basic BiDf-MKD aims to transfer the meta knowledge of a collection of APIs to one single meta model. A small number of APIs (e.g., 100 APIs) are insufficient to represent the actual underlying task distributions and makes it easy to overfit, leading to poor generalization ability for the new unseen tasks. 
To make our method work well within the setting of limited API budgets, we propose task memory replay to diversify the underlying task distribution by covering more interpolated tasks. 
We design a memory bank with the first-in-first-out structure to store the previous recovered task data from each API. 
We then generate new tasks that interpolate between the previous tasks.

Suppose the memory bank $\mathcal{B}$ has stored the recovered task data $\{\boldsymbol{S}_i,\boldsymbol{Q}_i\}_{i=0}^{T}$ recovered from the APIs $\{A_i\}_{i=0}^{T}$. Each task corresponds to a different label space. We generate a new task with a new label space by randomly resampling the class labels and the corresponding support set $(\boldsymbol{S}_m,\boldsymbol{Y}_{\boldsymbol{S}_m})$ and query set $(\boldsymbol{Q}_m,\boldsymbol{Y}_{\boldsymbol{Q}_m})$ from the memory bank. These new interpolated tasks do not correspond to any given API and thus diversify the task distribution, leading to better generalization to unseen tasks.
For these interpolated tasks, we adopt MAML \cite{finn2017model}, consistent with the bi-level structure of BiDf-MKD, to update the meta model by minimizing
\begin{equation}
\small
    \begin{split}
&\!\!\min_{\boldsymbol{\theta}}\ \mathcal{L}_{mem}^{outer}=\mathcal{L}_{cls}(F(\boldsymbol{Q}_m;\boldsymbol{\theta}_m),\boldsymbol{Y}_{\boldsymbol{Q}_m}), \\
        &\quad \rm{s.t. } \ \ \boldsymbol{\theta}_m=\min_{\theta}\mathcal{L}_{mem}^{inner}\triangleq        \min_{\theta}\mathcal{L}_{cls}(F(\boldsymbol{S}_m;\boldsymbol{\theta}),\boldsymbol{{Y}}_{\boldsymbol{S}_m}).
    \end{split}
    \label{eq:memory}
\end{equation}
The moment to perform task memory replay is flexible. For example, each API may come in a sequential way and we can perform task memory replay at the interval of two adjacent APIs or in exceptional cases where network interruption happens. Task memory replay does not need to query the APIs online, thus requiring no network connection and making our framework more stable. Overall, we integrate BiDf-MKD and task memory replay in an end-to-end manner, which is summarized in \cref{alg:blackboxDFML} of \cref{app:summarizedAlgorithhm}.
\section{Experiments}
We verify the effectiveness of our proposed BiDf-MKD framework in various real-world scenarios (API-SS, API-SH, and API-MH) with comprehensive ablation studies.

\subsection{Experimental Setup}
\label{subsec:experimentalSetup}
\textbf{Baselines.}\ \textbf{(i) Random}. Randomly parameterize the meta-initialization for meta testing. \textbf{(ii) Best-API.} We select the API with the highest reported accuracy to directly predict the query set during meta testing.  \textbf{(iii) Single-DFKD.} Single-level data-free knowledge distillation. Update the meta model in a sequential manner, with only single-level data-free knowledge distillation (\cref{eq:inner}). This baseline only transfers the task-specific knowledge sequentially instead of meta knowledge from APIs to meta model. 
\textbf{(iv) Distill-Avg.} We perform single-level data-free knowledge distillation (\cref{eq:inner}) for each API to obtain a surrogate white-box model. We average all surrogate model parameters layer-wise as the meta-initialization. \textbf{(v) White-box DFML.} Perform our BiDf-MKD in an ideal white-box setting, where the actual first-order (FO) gradients of the parameters inside the APIs are available for task recovery. This baseline provides an upper bound of performance for black-box DFML.

\begin{figure}[!t]
    \centering
    %\vspace{-0.2cm}
    \includegraphics[width=0.85\linewidth]{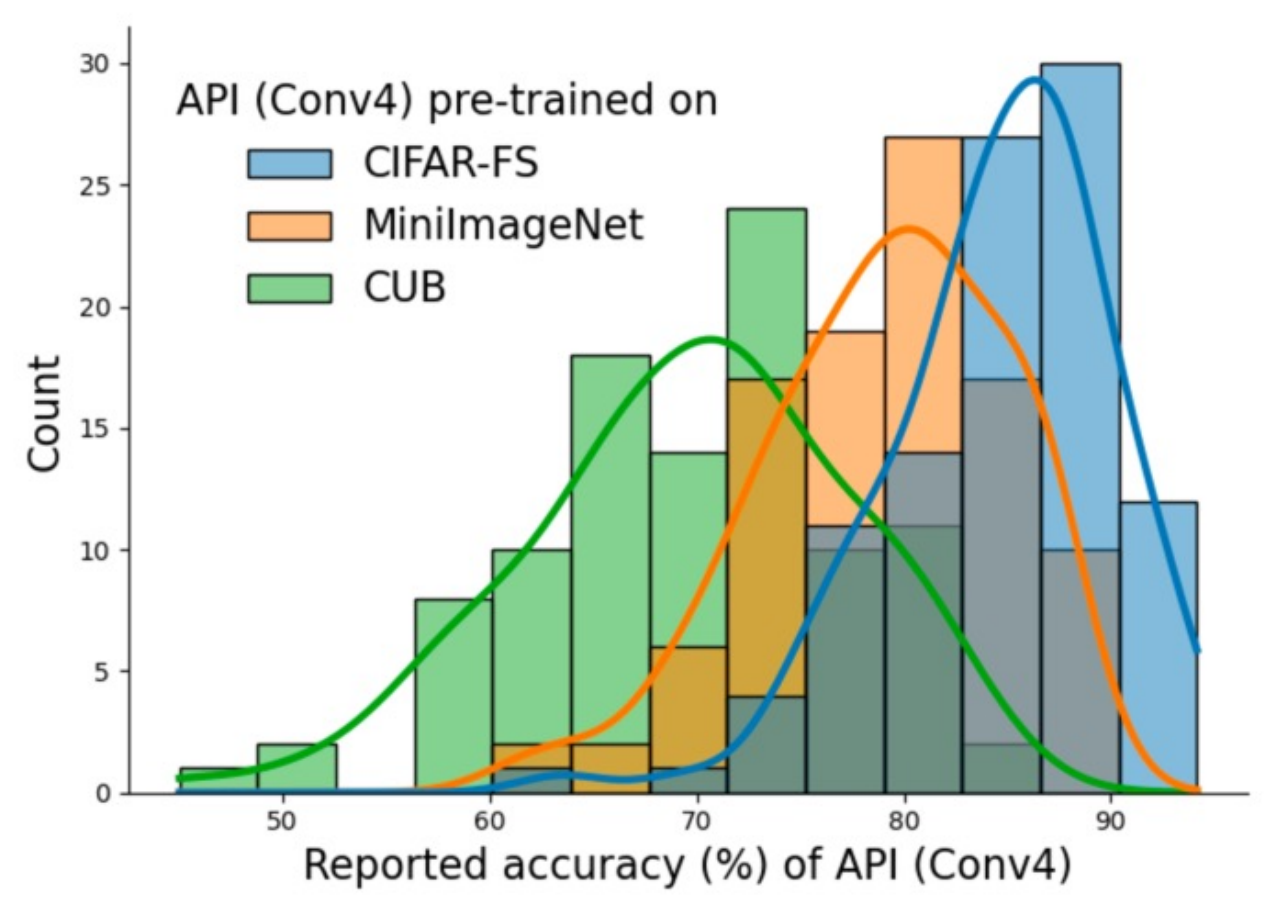}
    \vspace{-0.2cm}
    \caption{Histogram of the reported accuracy of APIs.}
    \vspace{-0.2cm}
    \label{fig:apiacc}
\end{figure}

\textbf{API quality.}\ \cref{fig:apiacc} shows the histogram of the reported accuracy of APIs (Conv4) pre-trained on CIFAR-FS, MiniImageNet and CUB, respectively. We provide more statistical results of APIs with other architectures (ResNet10 and ResNet18) in \cref{app:apiacc_all}. 
We argue that our framework should work well in the real-world scenario, where some APIs may be with relatively low accuracy. Refer to \cref{app:apiacc_all} for more discussions on robustness against the API quality variation.

\textbf{Implementation details \& Datasets.}\ We evaluate our BiDf-MKD framework on the meta-testing subsets of \textbf{CIFAR-FS} \cite{bertinetto2018meta}, \textbf{MiniImageNet} \cite{vinyals2016matching}, and \textbf{CUB-200-2011} (CUB) \cite{wah2011caltech}. Refer to \cref{app:dataset} and \cref{app:implementation} for detailed dataset setup and implementation details for three scenarios.

\subsection{Experiments of black-box DFML in API-SS}
\label{subsec:ss}

\textbf{Overview.}\ We first perform experiments in \textbf{API-SS} scenario, where all APIs are designed for solving different $5$-way tasks from the same meta training subset (CIFAR-FS, MiniImageNet or CUB) with the same architecture (Conv4).
% We take Conv4 as the meta model architecture. More implementation details is provided in \cref{app:implementation}.

\textbf{Results.}\ \cref{tab:ss} shows the results for 5-way classification in API-SS scenario. For 1-shot learning, ours outperforms the best baselines by $11.24\%$, $8.15\%$ and $8.09\%$ on three datasets, respectively. For 5-shot learning, ours outperforms the best baselines by $20.02\%$, $18,23\%$ and $21.46\%$ on three datasets, respectively. The results show that simply parameterizing the meta-initialization does not work for meta-learning. We observe a significant performance reduction of the best API from about $90\%$ to about $20\%$ because of the non-overlapping label space between the best API and the meta testing tasks and we can not fine-tune the API with the support 
set of new unseen tasks because of the black-box setting. The single-DFKD simply transfers the task-specific knowledge to the meta model in a sequential way; its bad performance reveals the accumulated task-specific knowledge is not beneficial to the efficient learning of new unseen tasks. Distill-Avg fuse all surrogate white-box models layer-wise and then fine-tune the meta-initialization; it also dose not perform well because those surrogate models train on different tasks, thus lacking precise correspondence among them. Ours performs the best because of the strong generalization ability of our transferred meta knowledge, which enables the efficient learning of new unseen tasks.
%\vspace{-0.5cm}
\begin{table}[htbp]
  \centering
  \small
   \caption{Compare to baselines in API-SS scenario.} 
   %\vspace{-0.2cm}
   \scalebox{0.8}{
  \begin{tabular}{clcc}
    \toprule
     \textbf{API-SS}&\textbf{Method}  & 1-shot &5-shot \\
    \midrule
    \multirow{5}{*}{\shortstack{\textbf{CIFAR-FS}\\\textbf{5-way}}} & Random & 20.35 $\pm$ 0.42 &  20.59 $\pm$ 0.45\\\
     & Best-API & 19.04 $\pm$ 0.68 &  19.04 $\pm$ 0.67 \\\
     & Single-DFKD & \ 20.04 $\pm$ 0.63 & 20.14 $\pm$ 0.64\\\
     & Distill-Avg & 24.24 $\pm$ 0.46 &  27.56 $\pm$ 0.51 \\
     & Ours & \textbf{ 35.48 $\pm$ 0.67} & \textbf{47.58 $\pm$ 0.74}\\
    \midrule
    \midrule
    \multirow{5}{*}{\shortstack{\textbf{MiniImageNet}\\\textbf{5-way}}} & Random & 21.20 $\pm$ 0.38 &  21.13 $\pm$ 0.37 \\
     & Best-API & 20.51 $\pm$ 0.63 & 20.39 $\pm$ 0.62 \\
     & Single-DFKD &  20.03 $\pm$ 0.60& 20.14 $\pm$ 0.66  \\
     & Distill-Avg & 20.53 $\pm$ 0.20 &  21.24 $\pm$ 0.24 \\
     & Ours & \textbf{ 29.35 $\pm$ 0.60} & \textbf{39.47 $\pm$ 0.64}\\
     \midrule
     \midrule
    \multirow{5}{*}{\shortstack{\textbf{CUB}\\\textbf{5-way}}} & Random &21.09 $\pm$ 0.38 &  21.11 $\pm$ 0.37\\
     & Best-API & 19.99 $\pm$ 0.69 &  19.95 $\pm$ 0.70  \\
     & Single-DFKD & 19.56 $\pm$ 0.64& 20.06 $\pm$ 0.64 \\
     & Distill-Avg & 21.07 $\pm$ 0.25 &  21.97 $\pm$ 0.30 \\
     & Ours & \textbf{29.10 $\pm$ 0.64} & \textbf{  43.43 $\pm$ 0.66}\\
     \bottomrule
  \end{tabular}}
  \label{tab:ss}
  \vspace{-0.2cm}
\end{table}

\subsection{Experiments of black-box DFML in SH}

\textbf{Overview.}\ We then perform experiments in a more realistic scenario, \textbf{API-SH}, where all APIs are designed for solving different tasks from the same meta training subset (CIFAR-FS, MiniImageNet or CUB) but with heterogeneous architectures (Conv4, ResNet10 and ResNet18).

\textbf{Results.}\ \cref{tab:sh} shows the result for 5-way classification in API-SH scenario. For 1-shot learning, ours outperforms the best baselines by $12.76\%$, $9.35\%$ and $9.02\%$ on three datasets, respectively. For 5-shot learning, ours outperforms the best baselines by $21.01\%$, $18.61\%$ and $22.87\%$ on three datasets, respectively. All baselines can not effectively solve the black-box DFML problem in API-SH scenario with the similar reasons of API-SS discussed in \cref{subsec:ss}. Ours is far better than all baselines and can apply to the API-SH scenario without any change because the meta knowledge distillation involved in our BiDf-MKD framework imposes no restriction on the underlying model architectures and scale inside each black-box API.

\begin{table}[htbp]
  \centering
  \small
   \caption{Compare to baselines in API-SH scenario.} 
   %\vspace{-0.2cm}
   \scalebox{0.8}{
  \begin{tabular}{clcc}
    \toprule
     \textbf{API-SH}&\textbf{Method}  & 1-shot &5-shot \\
    \midrule
    \multirow{5}{*}{\shortstack{\textbf{CIFAR-FS}\\\textbf{5-way}}} & Random & 20.35 $\pm$ 0.42 &  20.59 $\pm$ 0.45 \\
     & Best-API &  19.04 $\pm$ 0.68 &  19.04 $\pm$ 0.67\\
     & Single-DFKD &19.56 $\pm$ 0.67 & 20.06 $\pm$ 0.60\\
     & Distill-Avg & 22.82 $\pm$ 0.38 &  25.91 $\pm$ 0.45 \\
     & Ours & \textbf{35.58 $\pm$ 0.79} & \textbf{46.92 $\pm$ 0.77}\\
    \midrule
    \midrule
    \multirow{5}{*}{\shortstack{\textbf{MiniImageNet}\\\textbf{5-way}}} & Random &21.20 $\pm$ 0.38 &  21.13 $\pm$ 0.37\\
     & Best-API & 20.51 $\pm$ 0.63& 20.39 $\pm$ 0.62\\
     & Single-DFKD &20.11 $\pm$ 0.64&20.23  $\pm$ 0.66 \\
     & Distill-Avg & 20.32 $\pm$ 0.22 &  20.67 $\pm$ 0.24 \\
     & Ours & \textbf{30.55 $\pm$ 0.62} & \textbf{39.74 $\pm$ 0.65}\\
     \midrule
    \midrule
    \multirow{5}{*}{\shortstack{\textbf{CUB}\\\textbf{5-way}}} & Random &21.09 $\pm$ 0.38 &  21.11 $\pm$ 0.37\\
     & Best-API & 19.99 $\pm$ 0.69 &  19.95 $\pm$ 0.70 \\
     & Single-DFKD &20.13 $\pm$ 0.66&  20.24$\pm$ 0.64 \\
     & Distill-Avg & 20.46 $\pm$ 0.24 &  21.02 $\pm$ 0.26 \\
     & Ours & \textbf{30.11 $\pm$ 0.58} & \textbf{43.98 $\pm$ 0.64}\\
     \bottomrule
  \end{tabular}}
  \label{tab:sh}
  \vspace{-0.2cm}
\end{table}

\subsection{Experiments of black-box DFML in API-MH}

\textbf{Overview.}\ We further perform experiments in a more challenging \textbf{API-MH} scenario, where all APIs are designed for solving different tasks from multiple meta training subsets (CIFAR-FS, MiniImageNet and CUB) with heterogeneous architectures (Conv4, ResNet10 and ResNet18) inside. For meta testing, we evaluate the meta-learned meta-initialization on unseen tasks from CIFAR-FS, MiniImageNet and CUB simultaneously.

\textbf{Results.}\ \cref{tab:mh} shows the results for 5-way classification in API-MH scenario. Ours has significant performance advantages ($11.21\%$ and $17.13\%$ for 1-shot and 5-shot, respectively) compared with all other baselines, which shows the superiority and broad applicability of our BiDf-MKD framework to work well with black-box APIs from multiple datasets with heterogeneous architectures. 

\vspace{-0.5cm}
\begin{table}[htbp]
  \centering
  \small
   \caption{Compare to baselines in API-MH scenario.} 
   %\vspace{-0.2cm}
   \scalebox{0.8}{
  \begin{tabular}{clcc}
    \toprule
     \textbf{API-MH}&\textbf{Method}  & 1-shot &5-shot \\
    \midrule
    \multirow{5}{*}{\shortstack{\textbf{5-way} }} & Random &20.88 $\pm$ 0.39&  21.00 $\pm$ 0.40\\
     & Best-API &  19.44 $\pm$ 0.65&  19.64 $\pm$ 0.66\\
     & Single-DFKD & 19.04 $\pm$ 0.66 & 19.68 $\pm$ 0.64\\
     & Distill-Avg & 21.57 $\pm$ 0.25 &  23.11 $\pm$ 0.29 \\
     & Ours & \textbf{32.78 $\pm$ 0.60} & \textbf{40.24 $\pm$ 0.65}\\
     \bottomrule
  \end{tabular}
  }
  \label{tab:mh}
\end{table}

\subsection{Ablation Studies}
\label{sec:ablationStudies}
\textbf{Effectiveness of each component of our framework.}\ \cref{tab:ablation} analyzes the effectiveness of each component on CIFAR-FS in API-SS scenario. We first introduce the \textbf{vanilla} only performing meta-learning via task memory replay (\cref{sub:memory}). The vanilla still achieves a significant performance gain compared with the best baselines in \cref{tab:ss} by $8.89\%$ and $16.89\%$ for 1-shot and 5-shot learning, which hints the feasibility for meta-learning on synthetic data. By adding \textbf{BiDf-MKD} to transfer the meta knowledge, we observe a performance gain of $1.96\%$ and $2.49\%$. The reason is that BiDf-MKD provides a way to leverage richer supervision (i.e., the semantic class relationship in the soft-label prediction from those black-box APIs) instead of the only hard-label supervision from the synthetic data. We also observe an improvement ($0.89\%$ and $0.79\%$) from our \textbf{boundary} query set recovery technique (\cref{fig:boundary}), which verifies its effectiveness for alleviating the knowledge vanish issue. With all components, we achieve the best performance with a boosting improvement of $2.35\%$ and $3.13\%$, thus demonstrating the effectiveness of the joint schema.

\begin{table}[h]
  \centering
  \small
  \caption{Ablation studies on CIFAR-FS in API-SS scenario.}
    %\vspace{-0.6cm}
  \scalebox{0.8}{
  \begin{tabular}{ccccc}
    \toprule
    \small
    \multirow{2}{*}{\textbf{API-SS}} & \multicolumn{2}{c}{Component} & \multicolumn{2}{c}{Accuracy} \\
    \cmidrule(r){2-3}
    \cmidrule(r){4-5}
    &\textbf{BiDf-MKD}&\textbf{Boundary}&5-way 1-shot&5-way 5-shot\\
    \midrule
    \textbf{Vanilla}&&&33.13 $\pm$ 0.66&44.45 $\pm$ 0.76\\
    \midrule
    &\checkmark&&35.09 $\pm$ 0.71&46.94 $\pm$ 0.74\\
    &&\checkmark&34.02 $\pm$ 0.70 &45.24 $\pm$ 0.74\\
    \midrule
    \textbf{Ours}&\checkmark&\checkmark&\textbf{35.48 $\pm$ 0.67} & \textbf{47.58 $\pm$ 0.74}\\
     \bottomrule
  \end{tabular}}
  
  \label{tab:ablation}
\end{table}

\textbf{Effectiveness of zero-order gradient estimator.}\ 
% \cref{tab:zofo} provides an unfair comparison of our BiDf-MKD framework in black-box and white-box setting, respectively.
\cref{tab:zofo} provides an unfair comparison with an unfair baseline, white-box DFML, by performing our BiDf-MKD in an ideal white-box setting where the actual first-order (FO) gradients of the parameters inside the APIs are available for task recovery. The accuracy of white-box DFML serves as the upper bound for that of black-box DFML. The minor performance gaps on three datasets demonstrate our zero-order (ZO) gradient estimator can provide reasonable gradient estimation, which can achieve comparable meta-learning performance.
\begin{table}[htbp]
  \centering
  \small
   \caption{Effectiveness of zero-order gradient estimator. Grey: unfair comparison with white-box DFML.} 
   %\vspace{-0.2cm}
   \scalebox{0.8}{
  \begin{tabular}{cccc}
    \toprule
    % \multirow{2}{*}{} & \multirow{2}{*}{Method} & \multicolumn{2}{c}{5-way} \\
    % \cmidrule(r){3-4}
    %  &  & 1-shot & 5-shot \\
     \textbf{API-SS}&\textbf{Method}  & 1-shot &5-shot \\
    \midrule
    \multirow{2}{*}{\shortstack{\textbf{CIFAR-FS}\\\textbf{5-way}}} & \cellcolor{gray!40}FO & \cellcolor{gray!40} 37.66 $\pm$ 0.75 &  \cellcolor{gray!40} 51.16 $\pm$ 0.79\\
     & ZO &  35.48 $\pm$ 0.67&  47.58 $\pm$ 0.74\\
    \midrule
    \midrule
    \multirow{2}{*}{\shortstack{\textbf{MiniImageNet}\\\textbf{5-way}}} & \cellcolor{gray!40}FO &  \cellcolor{gray!40}30.66 $\pm$ 0.59 &   \cellcolor{gray!40} 42.30 $\pm$ 0.64\\
     & ZO &    29.35 $\pm$ 0.60 & 39.47 $\pm$ 0.64 \\
     % \cmidrule(r){2-4}
     % & \cellcolor{gray!40}MAML$^{\dag}$ & \cellcolor{gray!40}&\cellcolor{gray!40} \\
     \midrule
    \midrule
    \multirow{2}{*}{\shortstack{\textbf{CUB}\\\textbf{5-way}}} & \cellcolor{gray!40}FO &  \cellcolor{gray!40}31.62 $\pm$ 0.60 &   \cellcolor{gray!40} 44.32 $\pm$ 0.69\\
     & ZO &  29.10 $\pm$ 0.64 &   43.43 $\pm$ 0.66\\
     \bottomrule
  \end{tabular}}
  \label{tab:zofo}
\end{table}

\begin{figure}
    \centering
    \includegraphics[width=0.9\linewidth]{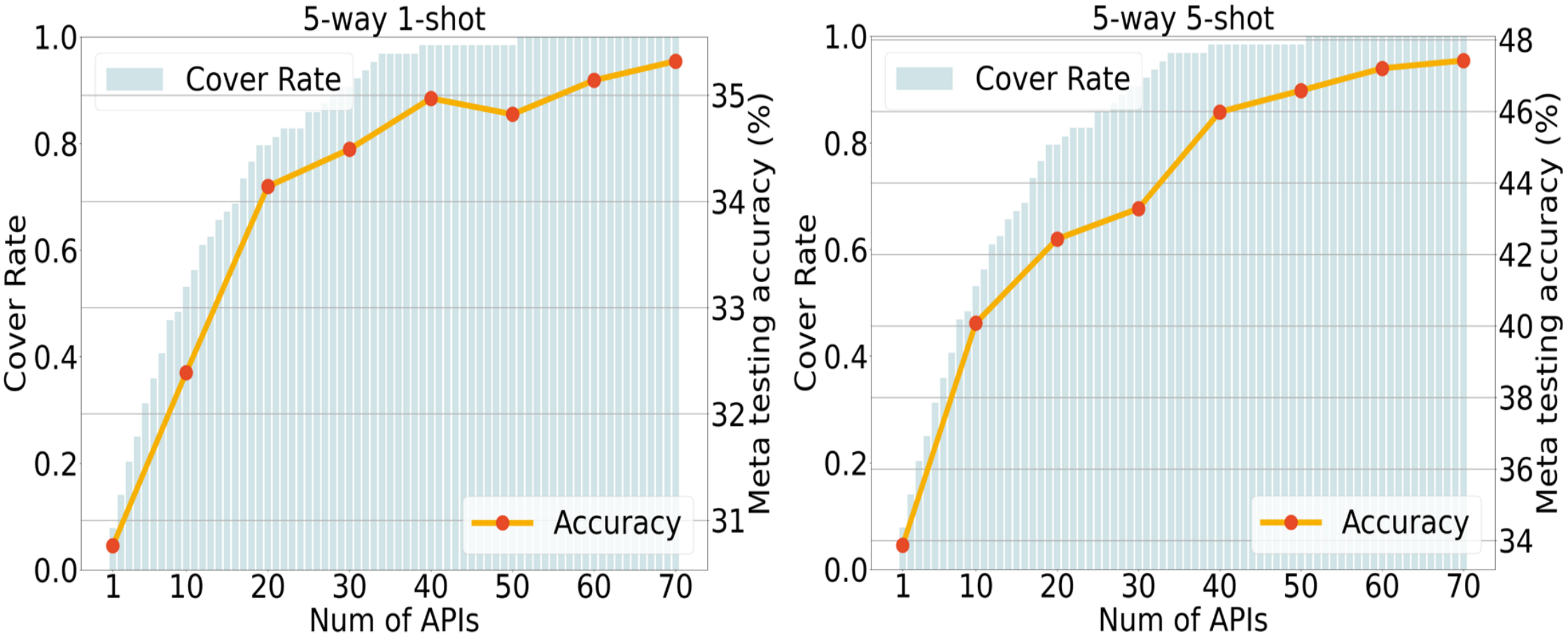}
    \vspace{-0.2cm}
    \caption{Effect of the number of APIs in API-SS scenario.}
    \label{fig:apinum}
    \vspace{-0.2cm}
\end{figure}

\textbf{Effect of the number of APIs.}\ \cref{fig:apinum} shows the performance difference with the different number of black-box APIs in API-SS scenario. Here, we further introduce an intrinsic factor, i.e., cover rate, which indicates the coverage rate of classes in the meta training subset. As the number of APIs increases, the cover rate increases, boosting the generalization ability for unseen tasks with higher meta testing accuracy. When the cover rate reaches $100\%$ (more than $50$ APIs), we can still observe a performance improvement because the additional APIs provide richer supervision of semantic relationships among different classes.

\textbf{Effect of the number of query times.}\ \cref{tab:numberquery} shows the effect of the number of query times (i.e., the value of $q$ in \cref{eq:estimation}) on the accuracy and time cost. More query times can lead to more accurate zero-order gradient estimation, thus leading to more accurate task recovery results and better meta-learning performance. Considering the time cost, we set $q=100$ in practice with comparable performance and tolerable time cost compared with the unfair performance bound of white-box DFML.

\begin{figure}
    \centering
    \includegraphics[width=0.95\linewidth]{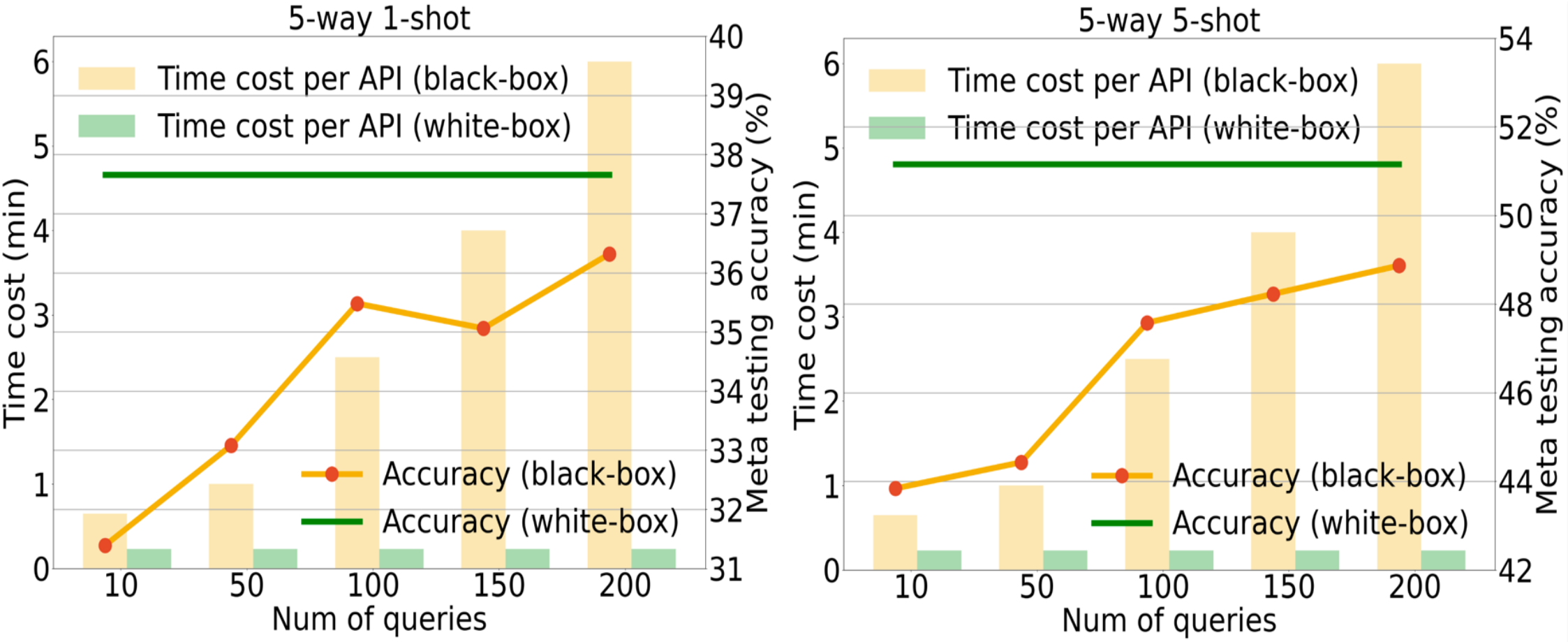}
    \vspace{-0.2cm}
    \caption{Effect of the number of query times on the accuracy and time cost. Here, white-box DFML provides unfair bounds of accuracy and time cost.}
    \vspace{-0.2cm}
    \label{fig:querynum}
\end{figure}

\begin{figure}
%\vspace{-0.2cm}
    \centering
    \includegraphics[width=0.95\linewidth]{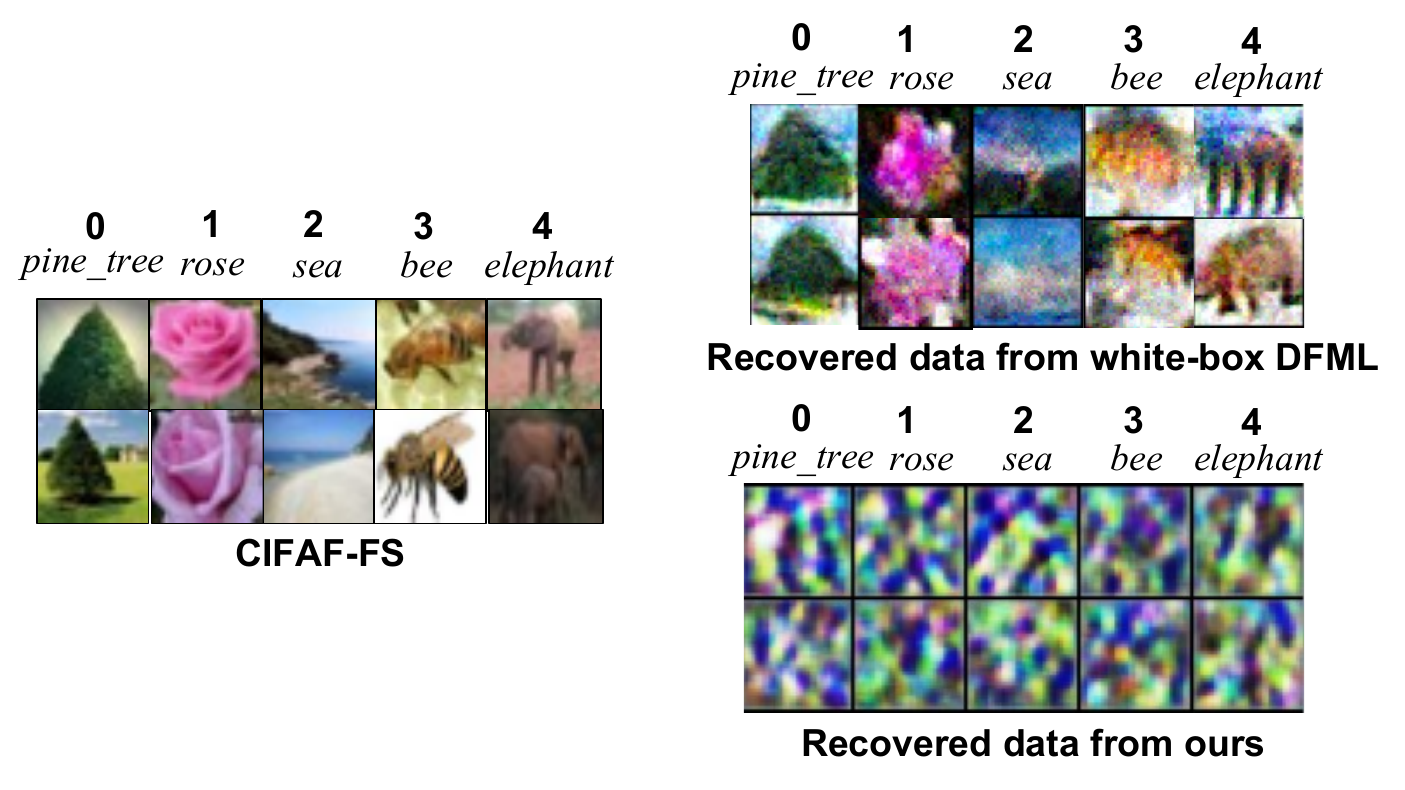}
    \vspace{-0.2cm}
    \caption{Visualization of recovered data on CIFAR-FS.}
    \vspace{-0.2cm}
    \label{fig:vix}
\end{figure}

\textbf{Data privacy.}\ A profound significance of black-box DFML is that we can achieve comparable meta-learning performance compared with the unfair white-box DFML without data privacy leakage. We argue that the real intention for releasing APIs (or pre-trained models) without data is to protect data privacy and security. However, as shown in \cref{fig:vix}, the recovered data from white-box DFML is highly similar to the original data, leaking sensitive data information and violating the original intentions. In contrast, ours recovers the data visually distinct from the original data, thus avoiding privacy leakage. Note that although the recovered data from ours look much different from the original data, ours still 
% outperforms the best baselines by a large margin (see \cref{tab:zofo}).
achieves a comparable meta-learning performance compared with the white-box DFML (see \cref{tab:zofo}).

% \begin{figure*}
%     \centering
%     \includegraphics[width=0.9\linewidth]{fig/vix_all.pdf}
%     \caption{***.}
%     \label{fig:vix_all}
% \end{figure*}

\section{Conclusion}
For the first time, we propose a practical and valuable setting of DFML, i.e., black-box DFML, which aims to meta-learn the meta-initialization from a collection of black-box APIs without access to the original training data and with only inference access, to enable efficient learning of new tasks without privacy leakage. 
To solve this challenging problem, we propose a novel BiDf-MKD framework integrated with task memory replay to transfer the general meta knowledge into one single model.
At last, we propose three real-world scenarios for a complete and practical evaluation of black-box DFML, where extensive experiments verify our framework's effectiveness and superiority.
\nocite{sun2022black,sun2022bbtv2}

\section*{Acknowledgements}
This work was supported by the National Key R\&D Program of China (2022YFB4701400/4701402), SZSTC Grant (JCYJ20190809172201639, WDZC20200820200655001), Shenzhen Key Laboratory (ZDSYS20210623092001004) and Beijing Key Lab of Networked Multimedia.

% \clearpage
\bibliography{ref}
\bibliographystyle{icml2023}

%%%%%%%%%%%%%%%%%%%%%%%%%%%%%%%%%%%%%%%%%%%%%%%%%%%%%%%%%%%%%%%%%%%%%%%%%%%%%%%
%%%%%%%%%%%%%%%%%%%%%%%%%%%%%%%%%%%%%%%%%%%%%%%%%%%%%%%%%%%%%%%%%%%%%%%%%%%%%%%
% APPENDIX
%%%%%%%%%%%%%%%%%%%%%%%%%%%%%%%%%%%%%%%%%%%%%%%%%%%%%%%%%%%%%%%%%%%%%%%%%%%%%%%
%%%%%%%%%%%%%%%%%%%%%%%%%%%%%%%%%%%%%%%%%%%%%%%%%%%%%%%%%%%%%%%%%%%%%%%%%%%%%%%

\clearpage
\appendix
\onecolumn
\vspace{\baselineskip}
\begin{center}
\LARGE
\textbf{Appendix}
\end{center}
% \tableofcontents
\section{Mutual Information and Entropy}
\label{app:mutualInformation}
\subsection{Mutual information}
Mutual information \cite{shannon1948mathematical,kreer1957question} $I({X}, {Y})$ measures the mutual dependence of two random variables ${X}$ and ${Y}$. Intuitively, it quantifies how much observing one random variable can reduce the uncertainty about the other random variable (i.e., uncertainty reduction) or measures how much observing one random variable can obtain the information about the other random variable (i.e., information gain). Below, we give the definition of discrete mutual information and continuous mutual information, respectively.

\begin{definition}
\textit{\textbf{Discrete} mutual information.} The  definition of mutual information $I(X, Y)$ of two discrete random variables ${X}$ and ${Y}$ is given by
\begin{equation*}
    I(X ; Y)=\sum_{y \in \mathcal{Y}} \sum_{x \in \mathcal{X}} P_{(X, Y)}(x, y) \log \frac{P_{(X, Y)}(x, y)}{P_{X}(x) P_{Y}(y)}.
\end{equation*}
\end{definition}
$X$ and $Y$ are random variables with the values over the space $\mathcal{X}$ and $\mathcal{Y}$, respectively. $P_{(X, Y)}$ is their joint distribution. $P_X$ and $P_Y$ are their marginal distributions. 

For two \textbf{continuous} random variables ${X}$ and ${Y}$, we replace the summations with the integrals.
\begin{definition}
\textit{\textbf{Continuous} mutual information.} The  definition of mutual information $I(X, Y)$ of two continuous random variables ${X}$ and ${Y}$ is given by
\begin{equation*}
    I(X ; Y)=\int_{\mathcal{Y}} \int_{\mathcal{X}} P_{(X, Y)}(x, y) \log \frac{P_{(X, Y)}(x, y)}{P_{X}(x) P_{Y}(y)}\ dx\ dy.
\end{equation*}
\end{definition}

\subsection{Entropy}

Entropy \cite{ben2008farewell} $H(X)$ is a measure of uncertainty of the random variable $X$. Below, we give the definition of discrete entropy and continuous entropy, respectively.
\begin{definition}
    \textit{\textbf{Discrete} entropy}. The definition of the entropy $H(X)$ of a discrete random variable $X$ is given by
    \begin{equation*}
        H(X)=-\sum_{x \in \mathcal{X}}{P_{X}(x) \log{P_{X}(x)}}.
    \end{equation*}
\end{definition}
We can also extend it to the \textbf{continuous} random variable by replacing the summations with the integrals.
\begin{definition}
    \textit{\textbf{Continuous} entropy}. The definition of the entropy $H(X)$ of a continuous random variable $X$ is given by
    \begin{equation*}
        H(X)=-\int_{\mathcal{X}}{P_{X}(x) \log{P_{X}(x)}}\ dx.
    \end{equation*}
\end{definition}

\subsection{Conditional entropy}

Conditional entropy quantifies the uncertainty of one random variable given that the value of the other random variable is known. Below, we give the definition of discrete conditional entropy and conditional continuous entropy, respectively.

\begin{definition}
    \textit{\textbf{Discrete} conditional entropy}. The definition of the conditional entropy $H(X | Y)$ of a discrete random variable $X$ given the other discrete random variable $Y$ is given by
    \begin{equation*}
        H(Y | X)=-\sum_{y \in \mathcal{Y}} \sum_{x \in \mathcal{X}} P_{(X, Y)}(x, y) \log{P_{(Y | X=x)}(y)}.
    \end{equation*}
\end{definition}

\begin{definition}
    \textit{\textbf{Continuous} conditional entropy}. The definition of the conditional entropy $H(X | Y)$ of a continuous random variable $X$ given the other continuous random variable $Y$ is given by
    \begin{equation*}
        H(Y | X)=-\int_{\mathcal{Y}} \int_{\mathcal{X}} P_{(X, Y)}(x, y) \log{P_{(Y | X=x)}(y)}\ dx\ dy.
    \end{equation*}
\end{definition}

\subsection{Relation between mutual information and entropy}
Here, we give a detailed deduction of the relation between mutual information and entropy for the case of discrete random variables $X$ and $Y$. The deduction for the case of continuous random variables is the same except for replacing the summations with the integrals.
\begin{equation*}
    \begin{aligned}
\mathrm{I}(X ; Y) & =\sum_{x \in \mathcal{X}, y \in \mathcal{Y}} P_{(X, Y)}(x, y) \log \frac{P_{(X, Y)}(x, y)}{P_X(x) P_Y(y)} \\
& =\sum_{x \in \mathcal{X}, y \in \mathcal{Y}} P_{(X, Y)}(x, y) \log \frac{p_{(X, Y)}(x, y)}{P_X(x)}-\sum_{x \in \mathcal{X}, y \in \mathcal{Y}} P_{(X, Y)}(x, y) \log P_Y(y) \\
& =\sum_{x \in \mathcal{X}, y \in \mathcal{Y}} P_X(x) P_{(Y \mid X=x)}(y) \log P_{Y \mid X=x}(y)-\sum_{x \in \mathcal{X}, y \in \mathcal{Y}} P_{(X, Y)}(x, y) \log P_Y(y) \\
& =\sum_{x \in \mathcal{X}} P_X(x)\left(\sum_{y \in \mathcal{Y}} P_{(Y \mid X=x)}(y) \log P_{(Y \mid X=x)}(y)\right)-\sum_{y \in \mathcal{Y}}\left(\sum_{x \in \mathcal{X}} P_{(X, Y)}(x, y)\right) \log P_Y(y) \\
& =-\sum_{x \in \mathcal{X}} P_X(x) \mathrm{H}(Y \mid X=x)-\sum_{y \in \mathcal{Y}} P_Y(y) \log P_Y(y) \\
& =-\mathrm{H}(Y \mid X)+\mathrm{H}(Y) \\
& =\mathrm{H}(Y)-\mathrm{H}(Y \mid X) .
\end{aligned}
\end{equation*}

\section{Full Architecture of Generator}

\cref{tab:sturcture_of_generator} lists the structure of the generator in our proposed BiDf-MKD framework. The generator takes the standard Gaussion noise as input and outputs the recovered data.  Here, $d_{\boldsymbol{z}}$ is dimension of Gaussian noise data $\boldsymbol{z}$, which is set as 256 in practice. 
The $negative\_slope$ of LeakyReLU is 0.2.
We set $img\_size$ as 32 for APIs pre-trained on CIFAR-FS and 84 for APIs pre-trained on MiniImageNet and CUB. We set the number of channels $nc$ as 3 for color image recovery and the number of convolutional filters $nf$ as 64.

\label{app:sturcture_of_generator}
\begin{table}[h]
    \centering
    \scalebox{0.7}{
    \begin{tabular}{ccc}
    \toprule
    \textbf{Notion} &\multicolumn{2}{c}{\textbf{Description}}\\
    \midrule
    $img\_size$ $\times$ $img\_size$&\multicolumn{2}{c}{ resolution of recovered image}\\
    $bs$ & \multicolumn{2}{c}{ batch size}\\
    $nc$ & \multicolumn{2}{c}{ number of channels of recovered image}\\
    $nf$ & \multicolumn{2}{c}{ number of convolutional filters}\\
    FC($\cdot$) & \multicolumn{2}{c}{ fully connected layer;}\\
    BN&\multicolumn{2}{c}{ batch normalization layer}\\
    Conv2D($input$,\ $output$,$filter\_size$,\ $stride$,\ $padding$) & \multicolumn{2}{c}{ convolutional layer}\\
    \toprule
    \multirow{2}{*}{\textbf{Structure}} & \multicolumn{2}{c}{\textbf{Dimension}}\\
    & \textbf{Before} & \textbf{After}\\
     \midrule
         $\boldsymbol{z} \in \mathbb{R}_{d_{\boldsymbol{z}}} \sim \mathcal{N}(\boldsymbol{0},\boldsymbol{1})$&--- & $\left[\ bs, d_{\boldsymbol{z}}\ \right]$  \\
        \midrule
        FC($\boldsymbol{z}$) & $\left[\ bs, \textcolor{red}{d_{\boldsymbol{z}}}\ \right] $ & $ \left[\ bs, \textcolor{red}{2 \times nf \times (img\_size//4) \times (img\_size//4)}\ \right]$\\
        \midrule
        Reshape & $\left[\ bs, \textcolor{red}{2 \times nf \times (img\_size//4) \times (img\_size//4)}\ \right] $ & $ \left[\ bs, \textcolor{red}{2 \times nf}, \textcolor{red}{(img\_size//4)}, \textcolor{red}{(img\_size//4)}\ \right]$\\
        \midrule
         BN & $ \left[\ bs, 2 \times nf, (img\_size//4), (img\_size//4)\ \right]$&$ \left[\ bs, 2 \times nf, (img\_size//4), (img\_size//4)\ \right]$\\
        Upsampling &$ \left[\ bs, 2 \times nf, (img\_size//\textcolor{red}{4}), (img\_size//\textcolor{red}{4}))\ \right]$ & $ \left[\ bs, 2 \times nf, (img\_size//\textcolor{red}{2}), (img\_size//\textcolor{red}{2}))\ \right]$\\
        \midrule
        Conv2D($2 \times nf,\ 2 \times nf,\ 3,\ 1,\ 1$) & $ \left[\ bs, 2 \times nf, (img\_size//2), (img\_size//2))\ \right]$ & $ \left[\ bs, 2 \times nf, (img\_size//2), (img\_size//2))\ \right]$\\
        BN, LeakyReLU & $ \left[\ bs, 2 \times nf, (img\_size//2), (img\_size//2))\ \right]$&$ \left[\ bs, 2 \times nf, (img\_size//2), (img\_size//2))\ \right]$ \\
        Upsampling& $ \left[\ bs, 2 \times nf, (img\_size//\textcolor{red}{2}), (img\_size//\textcolor{red}{2}))\ \right]$ & $ \left[\ bs, 2 \times nf, img\_size, img\_size\ \right]$\\
        \midrule
        Conv2D($2 \times nf,\ nf,\ 3,\ 1,\ 1$) & $ \left[\ bs, \textcolor{red}{2 \times nf}, img\_size, img\_size\ \right]$ & $ \left[\ bs, \textcolor{red}{nf}, img\_size, img\_size\ \right]$\\
        BN, LeakyReLU& $ \left[\ bs, nf, img\_size, img\_size\ \right]$ & $ \left[\ bs, nf, img\_size, img\_size\ \right]$\\
        Conv2D($nf,\ nc,\ 3,\ 1,\ 1$) & $ \left[\ bs, \textcolor{red}{nf}, img\_size, img\_size\ \right]$ & $ \left[\ bs, \textcolor{red}{nc}, img\_size, img\_size\ \right]$\\
        Sigmoid & $ \left[\ bs, nc, img\_size, img\_size\ \right]$& $ \left[\ bs, nc, img\_size, img\_size\ \right]$\\
        \bottomrule
    \end{tabular}
    }
    \caption{Detailed structure of generator in our proposed BiDf-MKD framework. We highlight the dimension change in \textcolor{red}{red}.}
    \label{tab:sturcture_of_generator}
\end{table}

\section{More Discussions}
\subsection{Discussions on robustness against API quality variation}
\begin{figure*}[h]
    \centering
    \includegraphics[width=0.95\linewidth]{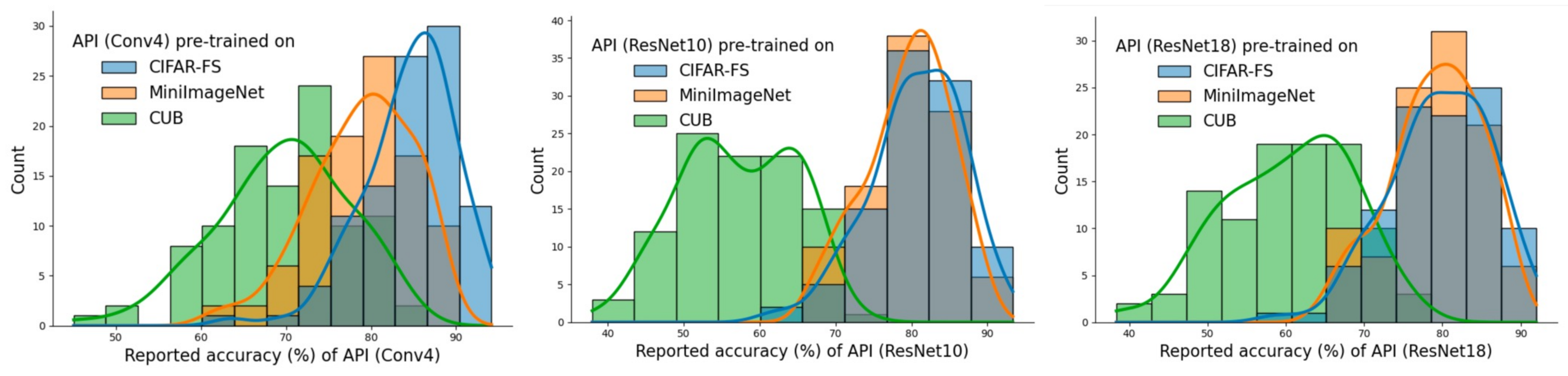}
    \vspace{-0.5cm}
    \caption{Histogram of reported accuracy of given APIs (Conv4, ResNet10, and ResNet18, respectively) pre-trained on CIFAR-FS, MiniImageNet, and CUB, respectively.}
    % \vspace{-0.5cm}
    \label{fig:apiacc_all}
\end{figure*}

\cref{fig:apiacc_all} shows the reported accuracy of given APIs (Conv4, ResNet10 and ResNet18, respectively) pre-trained on CIFAR-FS, MiniImageNet and CUB, respectively. We collect these black-box APIs to simulate the real-world scenario where various MaaS providers provide black-box APIs pre-trained on multiple datasets with heterogeneous architectures for solving different tasks. In addition, since we only have inference access to these black-box APIs without the original training data, we can not further improve the accuracy of these APIs by fine-tuning or re-training. In other words, our proposed framework should work well in a setting where a small number of APIs can be with relatively low accuracy. Therefore, as shown in \cref{fig:apiacc_all}, we perform our proposed BiDf-MKD method with the APIs of both high quality and relatively low quality. The results in \cref{tab:ss,tab:sh,tab:mh} show the superiority of our BiDf-MKD compared with the SOTA baselines against the API quality variation.

\label{app:apiacc_all}

% \subsection{Discussions on image resolution in API-MH scenario}
% In API-MH scenario, we are given a collection of black-box APIs designed for solving different tasks from multiple datasets with heterogeneous architectures inside. Note that each API may pre-train on images with different resolutions. For example, the APIs from CIFAR-FS may train on images of $32 \times 32$, while the APIs from MiniImageNet may train on images of $84 \times 84$. This can lead to confusion in task data recovery. Here, we provide two solutions: (i) we recover the images with a pre-defined resolution and resize the recovered images to the specific resolution required before feeding them to the APIs; (ii) we recover the images with the resolution required by each API and adopt the meta model architecture (e.g., adopt adaptive maxpooling technique before fully connected layer) to deal with varying input sizes. The former can obtain recovered images with the same resolution and the latter can obtain recovered images with different resolutions. In practice, we adopt the former because we can process the recovered images for the following bi-level meta knowledge distillation and task memory replay with unrestricted meta model architecture in a batch-wise way.

\subsection{Comparison between knowledge vanish issue and memorization issue of meta-learning}
The meta-learning memorization \cite{yin2019meta} describes an issue that the meta model overfits the training tasks, thus leading to poor generalization ability to unseen tasks. The memorization issue occurs the meta model can ignore the support set and directly perform well on the query set, i.e., the support set is useless.
 Our knowledge vanish issue is remarkably distinct from the memorization issue because the former occurs when the query set is useless. In addition, the knowledge vanish issue is more significant in the data-free meta-learning setting than in the data-based meta-learning setting because of the relatively low diversity of recovered data.

\section{Summarized Algorithm}
\label{app:summarizedAlgorithhm}
We summarize the algorithm of our end-to-end BiDf-MKD framework in \cref{alg:blackboxDFML}.

\begin{algorithm}[htbp]
\small
\DontPrintSemicolon
\SetKwInOut{Input}{Input}\SetKwInOut{Output}{Output}\SetKwInOut{Require}{Require}
\textbf{Input: }{Max iterations $N$; a collection of $T$ APIs $\{A_i\}_{i=0}^{T-1}$; the meta model $F(\cdot;\boldsymbol{\theta})$; memory bank $\mathcal{B}$; the batch size ${\rm BatchSize}$.
}\\
\textbf{Output: }{meta-initialization $\boldsymbol{\theta}$.}\\
\BlankLine
Randomly initialize $\boldsymbol{\theta}$\\
$\mathcal{B} \leftarrow [\ ]$\\
\For{$t \leftarrow 0$ \KwTo $N$}{
\For{$bs \leftarrow 0$ \KwTo ${\rm BatchSize}$}{
\If(\tcp*[f]{BiDf-MKD}){{\rm \textbf{no}} {\rm task memory replay}}{
Sample an API $A_i$\\
\tcp{support set recovery}
    Recover the support set $\boldsymbol{S}_i$ by minimizing \cref{eq:lossclsbatch} via zero-order gradient estimator\\
    \tcp{inner level of BiDf-MKD}
    Minimize \cref{eq:inner} via gradient descent w.r.t. to $\boldsymbol{\theta}$ to obtain the task-specific model parameters $\boldsymbol{\theta}_i$\\
    \tcp{boundary query set recovery}
    Recover the query set $\boldsymbol{Q}_i$ by minimizing \cref{eq:adaptive} via zero-order gradient estimator\\
    \tcp{outer level of BiDf-MKD}
    Minimize \cref{eq:outer} via gradient descent to update the meta model parameters $\boldsymbol{\theta}$\\
    $\mathcal{B} \leftarrow \mathcal{B} \cup \{\boldsymbol{S}_i, \boldsymbol{Q}_i\}$
}
\Else(\tcp*[f]{task memory replay}){
\tcp{task interpolation}
Construct interpolated task $\{\boldsymbol{S}_{m}, \boldsymbol{Q}_{m}\}$ from $\mathcal{B}$\\
\tcp{memory replay}
Minimize \cref{eq:memory} via gradient descent to update the meta model parameters $\boldsymbol{\theta}$\\
}
$bs \leftarrow bs+1$\\
}
$t \leftarrow t + 1$\\
}
\caption{Black-box data-free meta-learning.}
\label{alg:blackboxDFML}
\end{algorithm}

\section{Detailed Experimental Setup}
\subsection{Implementation details}
\label{app:implementation}

\textbf{API-SS.}\ For API-SS scenario, all APIs are designed for solving different tasks from the same meta training subset with the same model architecture inside. We take Conv4 as the architecture of the model inside each API and the meta model for API-SS. Conv4 is commonly used in meta-learning works \cite{finn2017model,chen2020variational,liu2020adaptive,wang2019simpleshot}, which consists of four convolutional blocks. Each block consists of $32\ 3 \times 3$ filters, a BatchNorm, a ReLU and a $2 \times 2$ max-pooling. 
All APIs are designed for solving different $5$-way tasks, which are constructed by randomly sampling $5$ classes from the same meta training subset (CIFAR-FS, MiniImageNet or CUB). We adopt Adam optimizer to pre-train the model inside each black-box API via standard supervised learning with a learning rate of $0.01$. In practice, we collect $100$ black-box APIs.
For BiDf-MKD, we recover $30$ images for the support set and query set, respectively. We adopt Adam optimizer to optimize the generator parameters $\boldsymbol{\theta}_{G}$ and input $\boldsymbol{z}$ simultaneously by minimizing \cref{eq:lossclsbatch} with the learning rate of $0.001$ for $200$ epochs. We adopt Adam optimizer to optimize the meta model parameters $\boldsymbol{\theta}$ by minimizing \cref{eq:outer} with the inner-level learning rate of $0.01$ and the outer-level learning rate of $0.001$.
For boundary query set recovery, we empirically set the coefficient $\lambda_Q$ as $1$.
For task memory replay, we adopt MAML to perform meta-learning on the interpolated tasks. 
We conduct MAML with the Adam optimizer with the inner-level learning rate of $0.01$ and the outer-level learning rate of $0.001$.
For the zero-order gradient estimator, we query each API with $100$ random direction vectors drawn from the sphere of a unit ball. We set the smoothing parameter $\mu$ as $0.005$ in \cref{eq:estimation}.

\textbf{API-SH.}\ For API-SH scenario, all APIs are designed for solving different tasks from the same meta training subset with heterogeneous model architectures inside. We take Conv4, ResNet-10 and ResNet-18 as the architectures inside the given APIs.
 Compared to Conv4, ResNet-10 and ResNet-18 are larger-scale neural networks. We take Conv4 as the meta model architecture. All APIs are designed for solving different $5$-way tasks, which are constructed by randomly sampling $5$ classes from the same meta training subset. 
 % All APIs are designed for solving different $5$-way tasks, which are constructed by randomly sampling $5$ classes from the same meta training subset (CIFAR-FS, MiniImageNet or CUB). We adopt Adam optimizer to pre-train the model inside each black-box API via standard supervised learning with a learning rate of $0.01$. In practice, we collect $100$ black-box APIs.
 The other configurations are the same as those of API-SS.

 \textbf{API-MH.}\ For API-MH scenario, all APIs are designed for solving different tasks from multiple meta training subsets with heterogeneous model architectures inside. We take Conv4, ResNet-10 and ResNet-18 as the architectures inside the given APIs.
  We take Conv4 as the meta model architecture. All APIs are designed for solving different $5$-way tasks, which are constructed by randomly sampling $5$ classes from multiple meta training subsets, including CIFAR-FS, MiniImageNet and CUB. For meta testing, we evaluate the meta-learned meta-initialization on unseen tasks from CIFAR-FS, MiniImageNet and CUB, respectively.
 The other configurations are the same as those of API-SS.

 \subsection{Datasets for Meta Testing}
\label{app:dataset}

\textbf{CIFAR-FS} \cite{bertinetto2018meta}, \textbf{MiniImageNet} \cite{vinyals2016matching} are commonly used in meta-learning, consisting of 100 classes with 600 images per class. We split each dataset into three subsets following \cite{wang2022metalearning}: 64 classes for meta training, 16 classes for meta validation and 20 classes for meta testing. 
In addition to these, we investigate \textbf{CUB-200-2011} (CUB) birds dataset \cite{wah2011caltech}, composing of 11,788 images of 200 bird species, to evaluate the effectiveness of our BiDf-MKD on fine-grained classification. We split into three subsets following \cite{chen2019closerfewshot}: 100 classes for meta training, 50 classes for meta validation and 50 classes for meta testing.
For CIFAR-FS, MiniImageNet and CUB, all splits are non-overlapping. 
Note that we have no access to the meta training subset in the DFML setting, and we only use meta testing subset for evaluation.

\subsection{Evaluation metric.}\ 
% Each $N$-way $K$-shot target task has a  support set  to adapt the meta-learned initialization to each target task. We take one-step gradient descent to perform fast adaptation in inner loop. 
We evaluate the performance by the average accuracy and standard deviation over 600 unseen target tasks sampled from meta testing subset. For API-SS and API-SH, we construct several meta testing tasks from one specific meta testing subset (i.e., CIFAR-FS, MiniImageNet or CUB), while for API-MH, we construct several meta testing tasks from all meta testing subsets (i.e., CIFAR-FS, MiniImageNet and CUB) equally.

\section{More Results}
\label{app:more_results}

\textbf{Effect of the number of APIs.}\ \cref{tab:numberapi} shows the effect of the number of black-box APIs on the meta testing accuracy. The results in \cref{tab:numberapi} are consistent with 
\cref{fig:apinum}. We additionally introduce an intrinsic factor, i.e., cover rate, which indicates the coverage rate of classes of meta training subset, to better illustrate the relationship between the number of APIs and the meta-learning performance. Refer to \cref{sec:ablationStudies} for detailed result analysis.
\begin{table}[h]
\vspace{-0.3cm}
  \centering
  \small
   \caption{Effect of the number of APIs in API-SS scenario.} 
   %\vspace{-0.2cm}
   \scalebox{0.99}{
  \begin{tabular}{cccc}
    \toprule
     \textbf{API-SS}&\textbf{APIs}  & 1-shot &5-shot \\
    \midrule
    \multirow{9}{*}{\shortstack{\textbf{CIFAR-FS}\\\textbf{5-way}}} 
     & 1 &  30.76 $\pm$ 0.64 & 33.87 $\pm$ 0.72\\
     & 10  &32.39 $\pm$ 0.66 & 40.08 $\pm$ 0.74\\
     & 20 & 34.14 $\pm$ 0.74 & 42.43 $\pm$ 0.74\\
     & 30 & 34.49 $\pm$ 0.66 & 43.28 $\pm$ 0.81\\
     & 40 & 34.97 $\pm$ 0.64 & 45.98 $\pm$ 0.76\\
     & 50 & 34.82 $\pm$ 0.71 & 46.58 $\pm$ 0.81\\
     & 60 & 35.14 $\pm$ 0.62 & 47.20 $\pm$ 0.74\\
     & 70 & 35.32 $\pm$ 0.64 & 47.42 $\pm$ 0.79\\
     & 100 & 35.48 $\pm$ 0.67 & 47.58 $\pm$ 0.74\\
     \bottomrule
  \end{tabular}
  }
  \label{tab:numberapi}
  %\vspace{-0.2cm}
\end{table}

\textbf{Effect of the number of queris.} \cref{tab:numberquery} shows the effect of the number of query times (i.e., the value of $q$ in \cref{eq:estimation}) on the meta testing accuracy. The results in \cref{tab:numberquery} are consistent with \cref{fig:querynum}. As we can see, more query times can lead to more accurate zero-order gradient estimation, thus leading to more accurate data recovery results and better meta-learning performance. Considering the time cost shown in \cref{fig:querynum}, we set $q=100$ in practice with comparable performance and tolerable time cost compared with the unfair performance bound of white-box DFML.

\begin{table}[h]
  \centering
  \small
   \caption{Effect of the number of queries in API-SS scenario.} 
   %\vspace{-0.2cm}
   \scalebox{0.99}{
  \begin{tabular}{cccc}
    \toprule
     \textbf{API-SS}&\textbf{Queries}  & 1-shot &5-shot \\
    \midrule
    \multirow{5}{*}{\shortstack{\textbf{CIFAR-FS}\\\textbf{5-way}}} 
    % & 5 &   $\pm$  &   $\pm$ \\
     & 10 &  31.39 $\pm$ 0.66 &  43.84 $\pm$ 0.74\\
     & 50 &  33.08 $\pm$ 0.68&   44.43 $\pm$ 0.74 \\
     & 100 & 35.48 $\pm$ 0.67 &  47.58 $\pm$ 0.74 \\
     & 150 & 35.06 $\pm$ 0.77 & 48.23 $\pm$ 0.81\\
     & 200 & 36.32 $\pm$ 0.77 & 48.87 $\pm$ 0.81\\
     \bottomrule
  \end{tabular}
  }
  \label{tab:numberquery}
  %\vspace{-0.2cm}
\end{table}

\textbf{Effect of the number of training classes for each task.}\ To evaluate the effect of the number of training class for each black-box API, we conduct the experiments in API-SS scenario, where each API trains for a 10-way classification problem. During meta-testing, we construct several 10-way meta testing tasks for evaluation. As shown in \cref{tab:dfmeta10way}, for CIFAR-FS, ours outperforms the best baseline by $9.45\%$ and $20.45\%$ for 1-shot learning and 5-shot learning, respectively. Compared to 5-way classification (\cref{tab:ss}), the meta testing accuracy of 10-way classification is
relatively lower because it is more challenging. Ours consistently outperforms all baselines in both 5-way and 10-way classification problems.

\begin{table}[htbp]
  \centering
  \small
   \caption{Effect of the number of training classes for each task.} 
   %\vspace{-0.2cm}
   \scalebox{0.99}{
  \begin{tabular}{clcc}
    \toprule
    % \multirow{2}{*}{} & \multirow{2}{*}{Method} & \multicolumn{2}{c}{5-way} \\
    % \cmidrule(r){3-4}
    %  &  & 1-shot & 5-shot \\
     \textbf{API-SS}&\textbf{Method}  & 1-shot &5-shot \\
    \midrule
    \multirow{5}{*}{\shortstack{\textbf{CIFAR-FS}\\\textbf{10-way}}} & Random & 10.12 $\pm$ 0.30 &  10.20 $\pm$ 0.26\\
     & Best-API &  9.86 $\pm$ 0.62& 9.94 $\pm$ 0.63\\
     & Single-DFKD & 10.02 $\pm$ 0.60 & 10.08 $\pm$ 0.64\\
     & Distill-Avg & 11.70 $\pm$ 0.21 &  12.22 $\pm$ 0.24 \\
     & Ours & \textbf{21.15 $\pm$ 0.37} & \textbf{32.67 $\pm$ 0.41}\\
     \bottomrule
  \end{tabular}}
  \label{tab:dfmeta10way}
  %\vspace{-0.2cm}
\end{table}

\textbf{Hyperparameter sensitivity.}\ In \cref{tab:sentivity}, we  evaluate the performance sensitivity of our BiDf-MKD framework for different values of $\lambda_{Q}$ ($0.1$, $1.0$ and $10.0$) in \cref{eq:adaptive}. The meta testing accuracy is stable and not sensitive to $\lambda_{Q}$ value variations for both 1-shot and 5-shot learning, which verifies the consistent superiority with different $\lambda_{Q}$ values. This advantage also makes it easy to apply our BiDf-MKD framework in practice.
\begin{table}[!h]
  \centering
  \small
  \caption{Hyperparameter sensitivity on  CIFAR-FS 5-way classification in API-SS scenario.}
    \label{tab:sentivity}
      \scalebox{0.99}{
  \begin{tabular}{ccc}
    \toprule

    $\lambda_{Q}$&   5-way 1-shot & 5-way 5-shot \\
    \midrule
    $\lambda_{Q}$ = 0.1 &35.32 $\pm$ 0.78 &47.72 $\pm$ 0.76\\
    $\lambda_{Q}$ = 1.0 & 35.48 $\pm$ 0.67 & 47.58 $\pm$ 0.74 \\
    $\lambda_{Q}$ = 10.0 & 35.06 $\pm$ 0.78 & 47.02$\pm$ 0.74\\
     \bottomrule
  \end{tabular}}
    %\vspace{-0.4cm}
\end{table}

\begin{figure}[!h]
    \centering
    \includegraphics[width=0.4\linewidth]{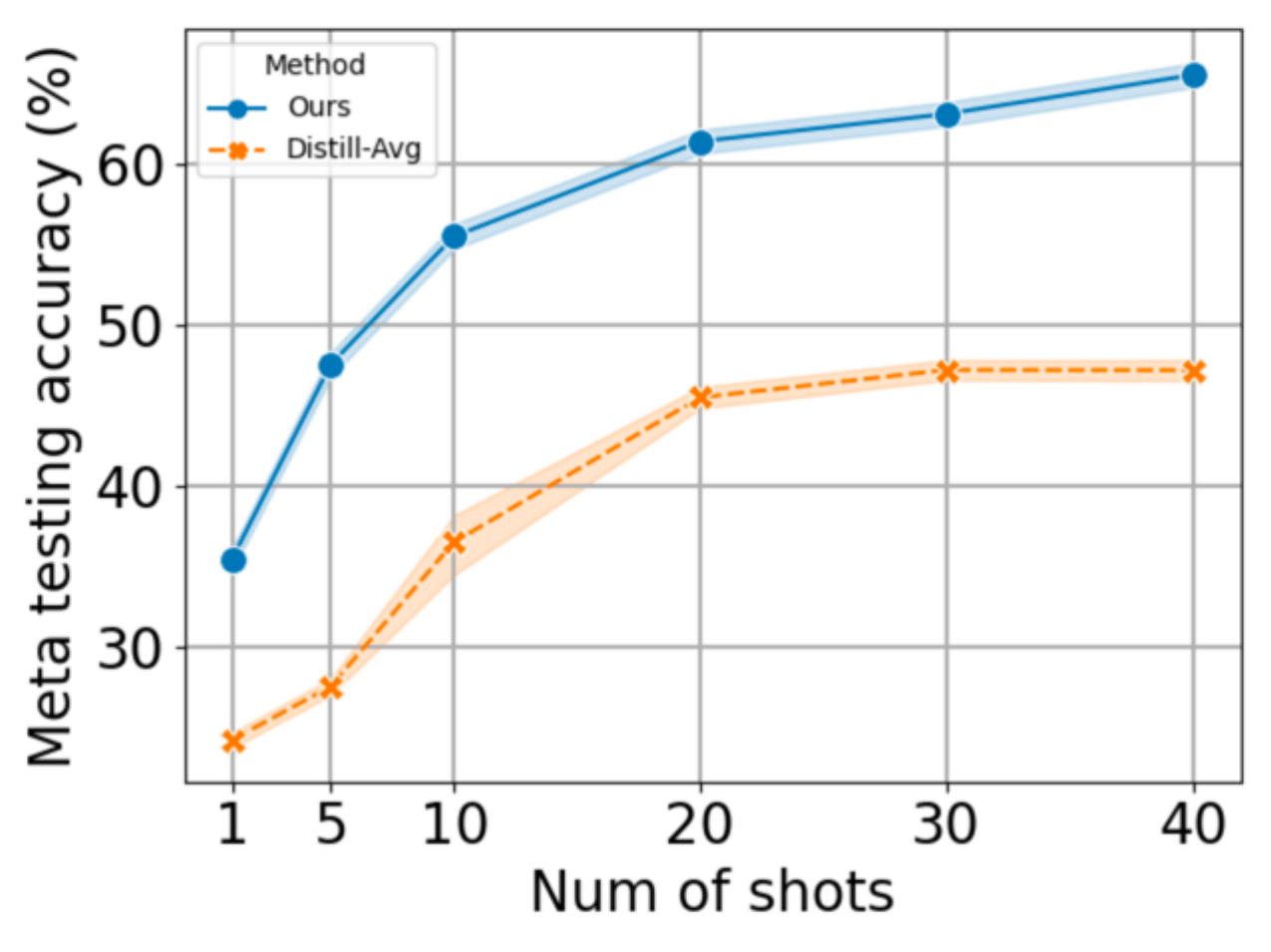}
    \vspace{-0.3cm}
    \caption{Larger-shot comparisons.}
    \label{fig:numshots}
    \vspace{-0.5cm}
\end{figure}

\textbf{Larger-shot comparisons.}\ We further investigate the meta-learning performance of our BiDf-MKD framework in the setting where there are more shots (i.e., the value of $K$) in the support set for the meta testing tasks (i.e., during meta-testing). In \cref{fig:numshots}, we conduct 5-way classification experiments on CIFAR-FS in API-SS scenario under different numbers of shots of meta testing tasks. We consider $K=\{1, 5, 10, 20, 30, 40\}$. Note that $K=\{1, 5\}$ is relatively larger than the common meta-learning setting where $K=\{10, 20, 30, 40\}$, but is smaller than the traditional supervised learning to train a complex neural network. For meta training, we set the same $K$ value for task memory replay. We consider the strong baseline ``Distill-Avg" discussed in \cref{subsec:experimentalSetup} considering its relatively better performance shown in \cref{tab:ss,tab:sh,tab:mh}. As shown in \cref{fig:numshots}, we can obtain a higher meta testing accuracy when $K$ increases for all methods and our BiDf-MKD framework outperforms the baseline at all the different shot settings. Simply averaging all surrogate models (i.e., Distill-Avg) lacking the meta-learning objective, leading to the bad performance in the low-shot setting (i.e., $K=\{1, 5\}$). Besides, Distill-Avg also does not perform well in the larger-shot setting (i.e., $K=\{10, 20, 30, 40\}$) because all surrogate models train on different tasks, thus lacking precise correspondence among them. In contrast, our BiDf-MKD framework outperforms the baseline at every $K$ value, showing its effectiveness across a broad spectrum of $K$ values.

%%%%%%%%%%%%%%%%%%%%%%%%%%%%%%%%%%%%%%%%%%%%%%%%%%%%%%%%%%%%%%%%%%%%%%%%%%%%%%%
%%%%%%%%%%%%%%%%%%%%%%%%%%%%%%%%%%%%%%%%%%%%%%%%%%%%%%%%%%%%%%%%%%%%%%%%%%%%%%%

\end{document}